\documentclass[10pt,twocolumn,letterpaper]{article}
\usepackage{cvpr}
\usepackage{multirow}
\usepackage{nicefrac}
\usepackage{mathtools}

\definecolor{cvprblue}{rgb}{0.21,0.49,0.74}
\usepackage[pagebackref,breaklinks,colorlinks,allcolors=cvprblue]{hyperref}
\usepackage{amsmath,amssymb}
\usepackage{color}
\newcommand*{\ShowNotes}{}
\ifdefined\ShowNotes
\newcommand{\colornote}[3]{{\color{#1}\bf{#2: #3}\normalfont}}
\else
  \newcommand{\colornote}[3]{}
\fi

\def\paperID{*****}
\def\confName{CVPR}
\def\confYear{2026}
\teaser{
    \centering
     \renewcommand{\arraystretch}{0.95}
    \begin{tabular}{c@{\hskip 3pt} c@{\hskip 3pt} c@{\hskip 3pt} c@{\hskip 3pt} c@{\hskip 3pt} c}
     \centering
       \multirow{3}{*}[+1ex]{\includegraphics[width=0.2\textwidth]{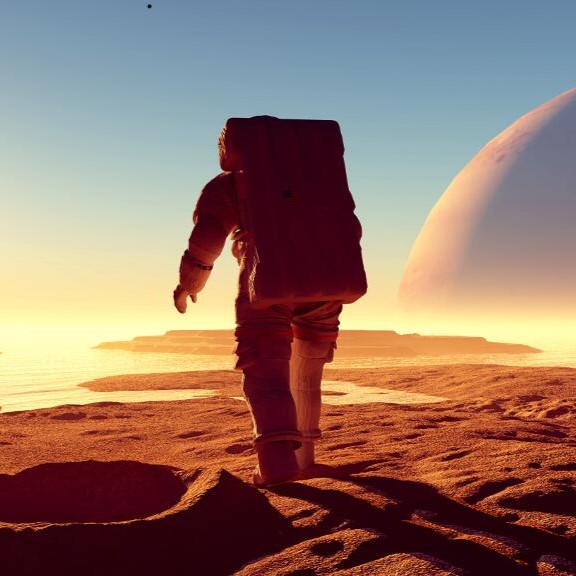}} &
      \includegraphics[width=0.11\textwidth]{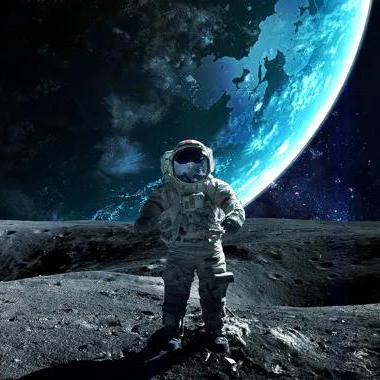} &  
      \includegraphics[width=0.11\textwidth]{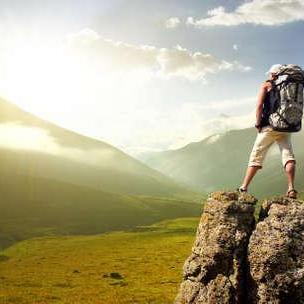} & 
      \includegraphics[width=0.11\textwidth]{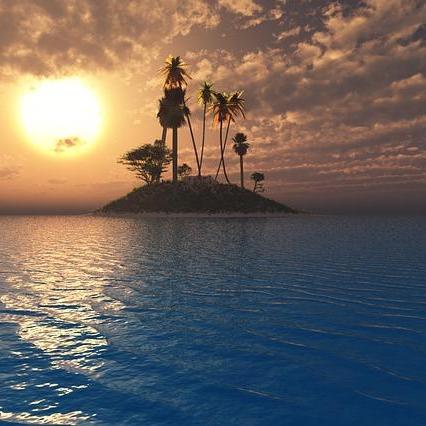} &
      \includegraphics[width=0.11\textwidth]{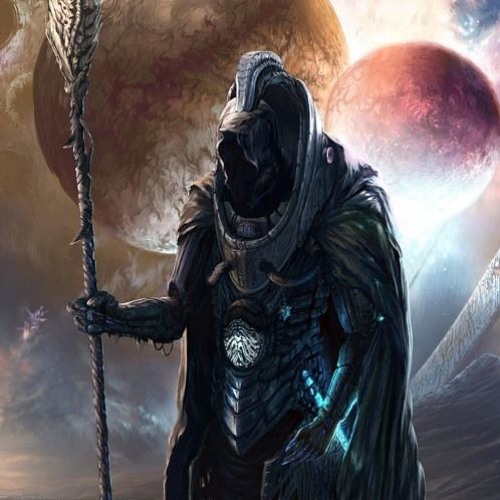} &
      \includegraphics[width=0.11\textwidth]{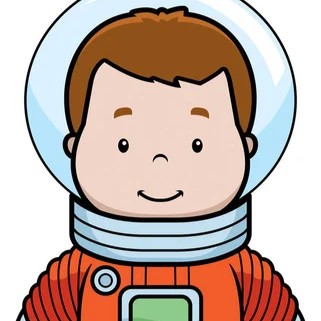}  \\
      &
      \includegraphics[width=0.11\textwidth]{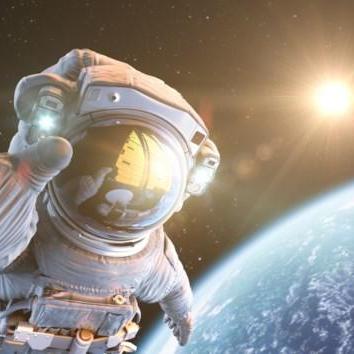} & 
      \includegraphics[width=0.11\textwidth]{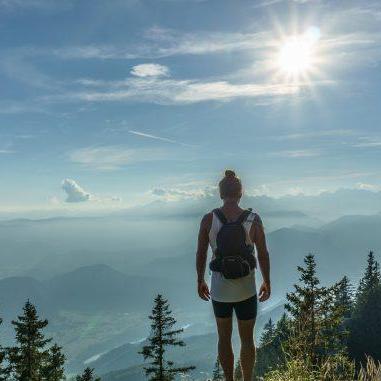} & 
      \includegraphics[width=0.11\textwidth]{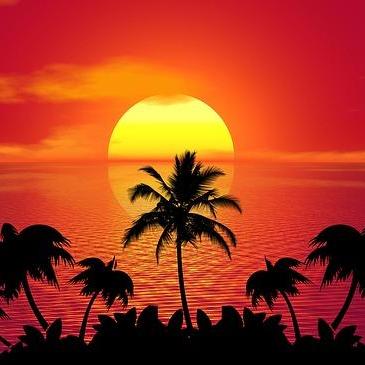} &
      \includegraphics[width=0.11\textwidth]{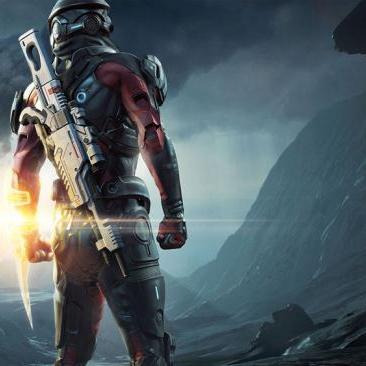} &
      \includegraphics[width=0.11\textwidth]{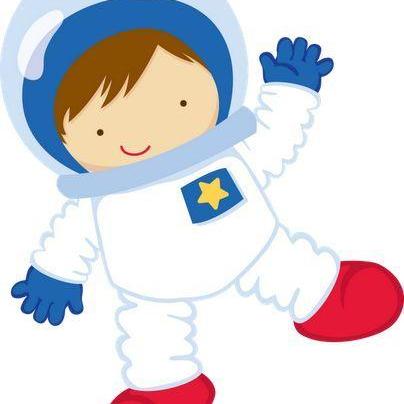} \\
      &
      \includegraphics[width=0.11\textwidth]{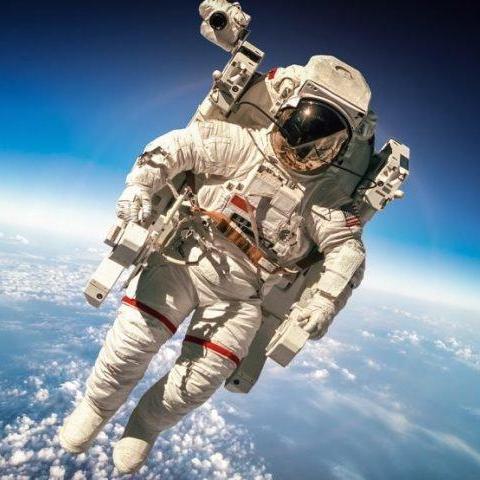} & 
      \includegraphics[width=0.11\textwidth]{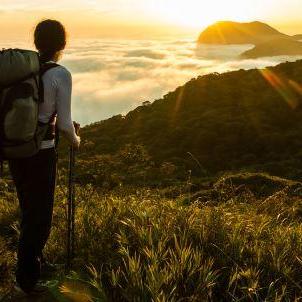} & 
      \includegraphics[width=0.11\textwidth]{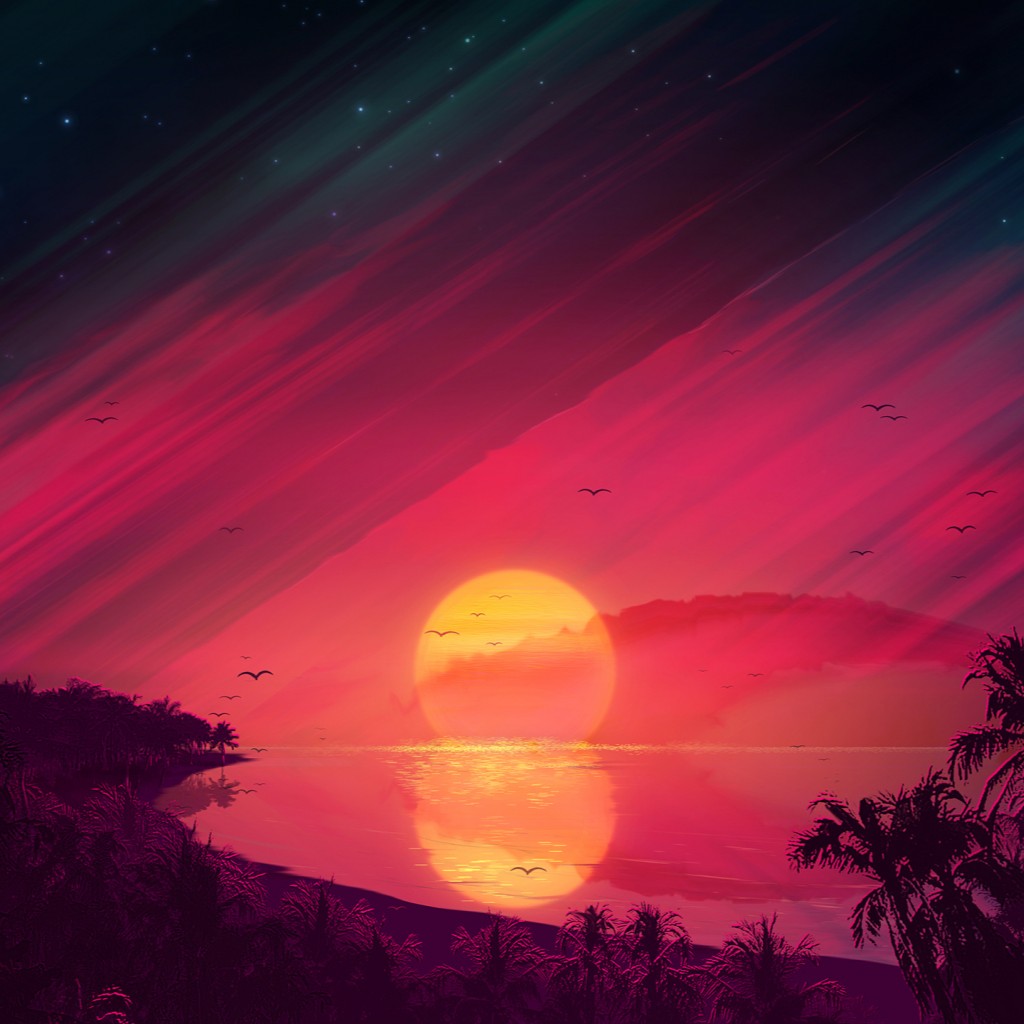} &
      \includegraphics[width=0.11\textwidth]{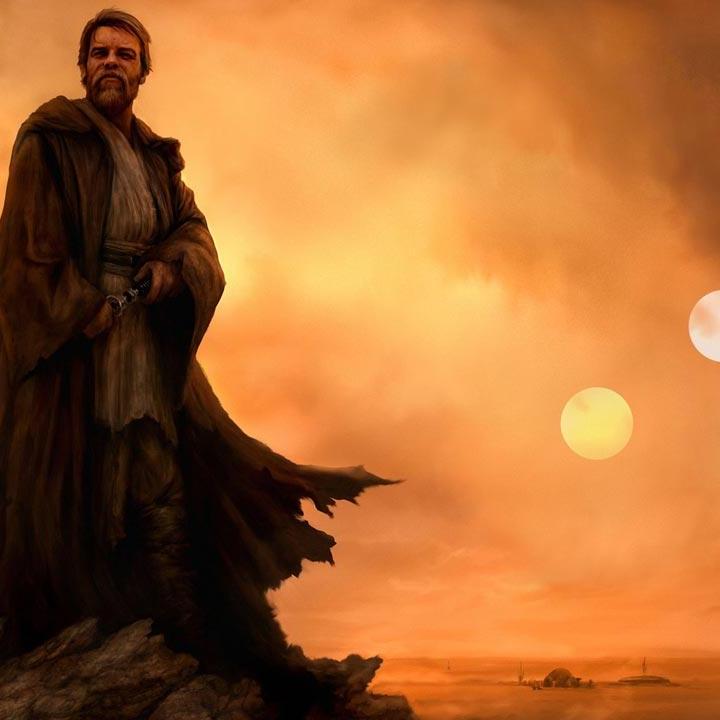} & 
      \includegraphics[width=0.11\textwidth]{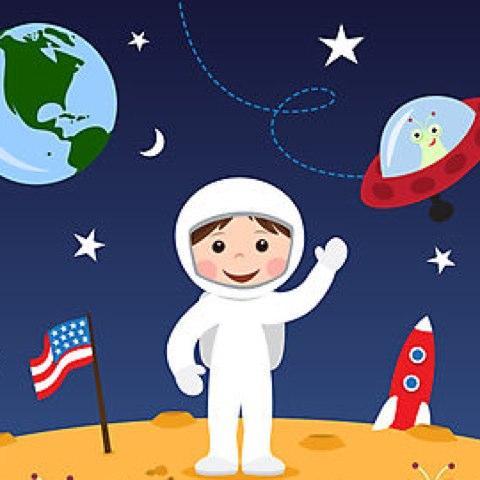}\\
        Input  & Concept 1 & Concept 2 & Concept 3 & Concept 4 & Concept 5 
\end{tabular}
\caption{
Given a query image (left) 
our method retrieves images from the dataset that share key concepts, relying exclusively on the visual characteristics of the input.
The extracted concepts can be interpreted as:
(1) Astronaut in space with a planet in the background
(2)~Human with a backpack on an exploration journey 
(3) Sunset
(4) Heroic figure in an otherworldly atmosphere
(5) Astronaut cartoon.}
\label{fig:Teaser}
}
\title{Concept Retrieval---What and How?}
\author{
Ori Nizan\\
  Technion, Israel\\
  {\tt\small snizori@campus.technion.ac.il}\\
  \and
  Oren Shrout\\
  Technion, Israel\\
  {\tt\small shrout.oren@gmail.com}\\
  \and
  Ayellet Tal\\
  Technion, Israel\\
  {\tt\small ayellet@ee.technion.ac.il}\\
}
\begin{document}
\maketitle
\begin{abstract}
A concept may reflect either a concrete or abstract idea.
Given an input image, this paper seeks to retrieve other images that share its central concepts, capturing aspects of the underlying narrative.
This goes beyond conventional retrieval or clustering methods, which emphasize visual or semantic similarity.
We formally define the problem, outline key requirements, and introduce appropriate evaluation metrics.
We propose a novel approach grounded in two key observations:
(1) While each neighbor in the embedding space typically shares at least one concept with the query, not all neighbors necessarily share the same concept with one another.
(2) Modeling this neighborhood with a bimodal Gaussian distribution uncovers meaningful structure that facilitates concept identification.
Qualitative, quantitative, and human evaluations confirm the effectiveness of our approach.
See the package on PyPI: https://pypi.org/project/coret/
\end{abstract}
\section{Introduction}
\label{sec:intro}
A concept is a mental representation that help humans categorize and interpret the world. 
In this sense, a concept is not merely a collection of visually similar features but rather an abstract, high-level grouping based on shared meaning, function, or context~\cite{margolis2007ontology}. 
Concepts can range from concrete (like "Forrest") to abstract (like "happiness") and are fundamental to thinking, reasoning, and learning.
In this paper, we introduce the task of retrieving images based on shared concepts.
This task can be seen as a generalization of image retrieval, which traditionally focuses on retrieving visually similar images from a dataset~\cite{dharani2013survey_content_based_image_retrieval, qazanfari2023advancements_in_content_based_image_retrieval}.
In contrast, our task emphasizes high-level semantics over visual similarity and aims to capture abstract and contextual meaning effectively.
For example, given an image of an astronaut exploring space (Fig.~\ref{fig:Teaser}), concept retrieval may return images of astronauts in various atmospheres and planets (Concept 1), exploratory journey (Concept 2), and so on.
In this example, the concept of an 'exploration journey' is better captured through activity and context rather than mere visual similarity.
This capability is especially valuable for AI-driven creative applications.
For example, in advertising, brands often seek images that express concepts such as innovation, exploration, or family values, rather than depicting a specific object.
In the arts and creative industries, artists and curators frequently retrieve works based on themes or concepts rather than literal objects.
In psychology, such retrieval can support studies of visual metaphors and the ways in which people interpret abstract ideas through images.
A key consideration is defining the essential requirements of the task. We identify four such requirements:
(1) {\em Relevance}: the retrieved images should reflect a concept present in the input image;
(2) {\em Consistency}: images retrieved for a particular concept should consistently represent that concept;
(3) {\em Inner-concept diversity}: images within a single concept should vary  rather than appear nearly identical;
(4) {\em Cross-concept diversity}: images retrieved for one concept should be semantically different from those of other concepts.
Once the requirements are defined, the next step is to develop an effective approach to address the problem.
Our method is based on the following observations. 
Each image in the embedding space neighborhood of a query tends to share at least one concept with it. 
At the same time, not all neighboring images necessarily share the same concept with one another.
However, if a sufficient number of neighbors express a particular concept, that concept can be reliably identified as a primary concept of the query.
The key challenge is to effectively leverage these relationships within the embedding space to extract meaningful concepts.
To isolate a concept embedding, we aim to partition the embeddings in the local neighborhood into two subsets: one containing images that represent the concept and another excluding it.
The challenge lies in achieving this without prior knowledge of the concept itself.
The key idea is to identify a {\em surrogate} embedding within the neighborhood, which will
 allow us to model the similarity distribution of neighboring embeddings as a bimodal Gaussian.
In this distribution, one Gaussian corresponds to the concept, while the other represents images that do not share the concept.
Since each image contains multiple concepts, once images sharing a particular concept are identified, the dataset embeddings should be updated to reduce that concept’s influence. Following this adjustment, the search for a new concept resumes. As the embeddings change, the input image’s neighborhood also shifts, allowing new concepts to emerge. This iterative process leverages the evolving embedding space to ensure diverse and relevant concepts.
But how should the results be evaluated? As this is a new task, dedicated evaluation metrics are needed.
We introduce four evaluation metrics, each corresponding to one requirement.
It is important to note that these requirements may not always align; for example, consistency and inner-concept diversity can sometimes be in conflict. Therefore, measuring each requirement separately is crucial, allowing applications to determine the appropriate balance based on their specific needs.
Additionally, a human evaluation methodology is proposed to capture subjective opinions on how well the results adhere to each requirement.
This paper makes the following contributions:
\begin{enumerate}
\item 
Defining a new problem that generalizes the image retrieval task, along with establishing its key requirements.
\item
Proposing a novel approach to address the task, which is both efficient and scalable.
Quantitative, qualitative, and human evaluation results demonstrate the method's effectiveness across heterogeneous datasets.
\item 
Introducing new evaluation metrics designed to assess these requirements effectively.
\end{enumerate}
\section{Related Work}
\label{sec:related}
This paper aims to retrieve images that share the underlying concepts as a given image.
The term {\em concept} has been interpreted in various ways in computer vision, often diverging from its psychological definition, as a mental representation that forms abstract, high-level groupings based on shared meaning, function, or context.
Within computer vision, our work can be viewed as a generalization of image retrieval.

\vspace{0.02in}
\noindent
{\bf Concepts in Computer Vision.}
The term "concept" has been interpreted in various ways in computer vision.
In most cases, it refers to an object or a style and has been applied in tasks such as image generation and editing~\cite{gal2022image,gal2023encoder,jia2023taming,jia2013visual,kumari2023multi,lee2023language,safaee2024clic}.
In Concept Bottleneck Models (CBMs)~\cite{koh2020concept_bottleneck_models,shang2024incremental_concept_bottleneck_models}, object-based concepts have been used to enhance explainability by learning interpretable representations that improve the transparency of decision-making.
These approaches can be broadly classified into:
(1) Language-guided extraction~\cite{andreas2017learning,oikarinen2023labelfreeconceptbottlenackmodels,yang2023language}, which leverages textual descriptions to define concepts.
(2) Vision-guided extraction~\cite{chen2019looks,chen2020concept,majellaro2024explicitlydisentangledinobjectcentriclearning,nauta2021neuralprototypetreesforinterpretable}, which derives concepts directly from images without relying on textual supervision. 
Other works have explored concepts for explainability.
In particular, Ghorbani et al.~\cite{ghorbani2019towards_automatic_concept_based_explanations} define concepts as image segments. For example, a "wheel" of a vehicle is considered a distinct concept.
In Chattopadhyay et al.~\cite{chattopadhyay2024information_maximization}, concepts are represented as sets of words from a predefined dictionary.
Our approach differs from the above by using the term concept in a broader sense, relying solely on images as input, without predefined linguistic guidance or a fixed vocabulary, and by extracting multiple concepts per image.

\vspace{0.02in}
\noindent
{\bf Content-based image retrieval.}
This is one of the classical tasks in computer vision.
The goal is to identify and return the most relevant images from a database based on specific visual features or content.
The classical methods retrieve the most similar images by comparing their feature vectors~\cite{
fu2016content_based_image_retrieval_based_on_CNN_and_SVM,
kashid2024optimizing_Content_Based_Image_Retrieval_System_Using_Convolutional_Neural_Network_Models,
li2020unsupervised,
rani2025efficient_content_based_image_retrieval_framework_using_separable_CNNs,
rastegar2023designinga_new_deep_convolutional_neural_network_for_content-based_image_retrieval_with_relevance_feedback,
rian2019content_based_image_retrieval_using_convolutional_neural_networks}. 
Recent approaches employ multimodal retrieval, where text and image features are jointly learned~\cite{chen2020uniter_universal_image_text_representation_learning, huo2021wenlan_bridging_vision_and_language_by_large_scale_multi_modal_pre_training,jia2021scaling_up_visual_and_vision_language_representation_learning_with_noisy_text_supervision} leveraging vision-language models (VLMs)~\cite{dfn_fang2023data, li23-blip2, radford2021clip}.
We refer the readers to recent surveys that highlight the evolution of retrieval methods~\cite{datta2008image,cao2022image_text_retrival, chen2021deep_image_retrival_a_survey, dubey2021decade_survey_of_content_based_image_retrieval_using_deep_learning, wan2014deep_learning_for_Content_Based_Image_retrieval_Comprehensive_study}. 
Similarly, our objective is to retrieve images that share common traits with a given input image.
However, we seek images that share multiple semantically concepts rather than  a single visual similar neighborhood.
Our concepts are discovered post-hoc from an existing embedding space, without language labels, attributes, or detectors.
\section{Method}
\label{sec:model}
Given an image, our  aim is to retrieve images that share the same concept(s).
The proposed method is designed to meet four key requirements:
\begin{enumerate}
     \item Relevance:
     Each retrieved image should contain a concept that is present in the input image.
   \item Consistency:
  Images retrieved for a given concept should accurately represent that concept.
   \item Inner-concept diversity:
   The retrieved images for a given concept should exhibit variation.
    \item Cross-concept diversity:
    The set of images retrieved for concept $i$ should visually and semantically differ from those retrieved for all previous concepts.
\end{enumerate}
Our approach is built on two key ideas:
(1) {\em Bimodal neighborhood structure:} Within the neighborhood of a given embedding, some images share a specific concept while others do not. 
This distinction can be modeled as a bimodal Gaussian distribution, though the exact concept remains unknown.
(2) {\em Concept surrogates:} The similarities of certain embeddings to their neighbors clearly exhibit a bimodal Gaussian structure.
These embeddings may serve as concept surrogates.
By analyzing the neighborhoods of these surrogates (within the query neighborhood), our method isolates the most prominent concepts in the image, consistent with the dataset.
The central question, then, is how to analyze this structure.
Our method operates in 4 steps:
(1) Identify the neighbors of the query embedding.
(2) Within this neighborhood, determine the most suitable surrogate neighbor.
(3) Extract images that share the concept common to both the surrogate neighbor and the query.
(4) Update the dataset and repeat from Step 2 to extract the next concept.
We elaborate below.

\vspace{0.02in}
\noindent
{\bf 1. Identify the query neighbors.}
Images located near each other in the embedding space are likely to share an underlying visual concept.
Therefore, given a query image, we begin by identifying its set of neighboring images, ensuring the {\em relevance} requirement is met.
We use cosine similarity:
\begin{equation} 
\text{Sim}(\mathbf{e}_i, \mathbf{e}_j) = \frac{\mathbf{e}_i \cdot \mathbf{e}_j}{|\mathbf{e}_i| |\mathbf{e}_j|}, \label{eq:sim}
\end{equation}
where $\mathbf{e}_i$ and $\mathbf{e}_j$ denote the embeddings of two images, $I_i$ and $I_j$.
To ensure that relevant neighbors are captured, we adopt a relatively large neighborhood size. This allows us to include images that share a common concept but may lie farther apart due to the influence of diverse attributes.
In our implementation, proximity is defined using a threshold $T = \sigma$, computed relative to the mean embedding distance $\mu$ of the dataset.
Thus, the neighborhood of the input embedding $e$ in the embedding space $\mathcal{D}$ is defined as follows:
\begin{equation}
{Neighbrhood}_{T}(\mathbf{e})=\{\mathbf{e}_j \mid \text{Sim}(\mathbf{e}, \mathbf{e}_j) \geq T, \mathbf{e}_j \in \mathcal{D}\}.
\label{eq:neighbr}
\end{equation}

\begin{figure}
    \centering
    \includegraphics[width=0.47\textwidth]{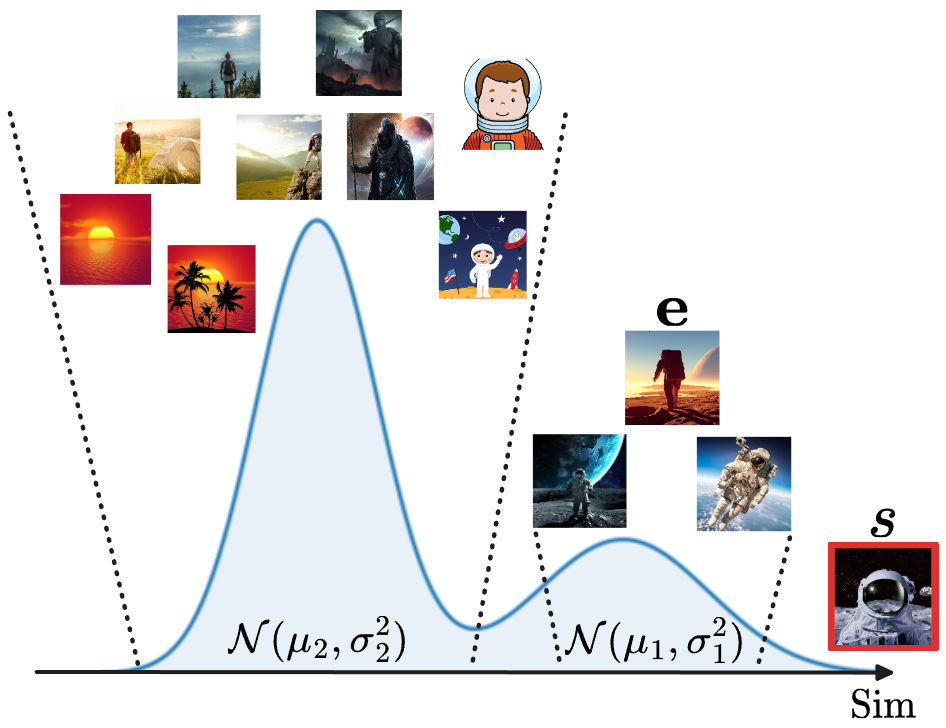} 
    \caption{
    {\bf Similarity score distribution.}
    This image shows a bimodal Gaussian of similarity scores between a surrogate \(\mathbf{s}\) and the input's neighbors from Fig.~\ref{fig:Teaser}.
The smaller (right) mode defines the 'concept' set; the larger (left) defines the 'non-concept' set.\\
}
    \label{fig:gmm}
\end{figure}

\noindent
{\bf 2. Determine a surrogate neighbor and its concept set.}
The goal of this step is to select an effective surrogate image from the neighborhood. This surrogate image helps retrieve a subset of images that share a common concept, thereby ensuring the {\em consistency} requirement is met.
To achieve this, we first compute pairwise similarities between all image embeddings 
$e_i$ and $e_j$ in the neighborhood.
We then construct a similarity-based histogram for each image in the set.
Recall our observation that certain embeddings 
exhibit a clear bimodal Gaussian distribution in their similarity values, making them suitable surrogates~$s$ for identifying a concept.
Formally, we apply Gaussian Mixture Modeling (GMM) to each histogram:
\begin{equation}
GMM(\text{Sim}(s, e)) =\ \pi_1 \mathcal{N}(\mu_1, \sigma_1^2) + \pi_2 \mathcal{N}(\mu_2, \sigma_2^2) ,
\label{eq:gmm}
\end{equation}
where $\mu_i$ and $\sigma_i$ are the means and standard deviations of the two Gaussians and $\pi_i$ are the mixture coefficients satisfying \( \pi_1 + \pi_2 = 1 \).
If two well-separated Gaussian distributions emerge, this indicates a potential separation between images that contain a certain concept and those that do not.
Figure~\ref{fig:gmm} illustrates the Gaussian distribution corresponding to the input image $e$, presented in Fig.~\ref{fig:Teaser}, and its surrogate $s$.
Next, to choose the concept from the multiple candidates, we apply the following two criteria to each GMM:
(1)~The number of samples in each Gaussian must exceed a threshold. This ensures that both sets are sufficiently representative of the data.
(2)~The input image must belong to the right Gaussian. 
This guarantees that the identified subset focuses on a concept that is present in the input image.
We rank the candidate surrogates to select the image that best separates the two Gaussian distributions, ensuring a clear distinction between the subsets.
This step directly supports inner-concept {\em consistency,} as a greater separation between the Gaussians indicates more distinct distributions, leading to a more consistent concept.
To quantify this separation, we define a separation metric score:  
\begin{equation}
  SepScore(s) = (\mu_1-\sigma_1)-(\mu_2+\sigma_2),  
\label{eq:score}
\end{equation}
where \( \mu_1 - \sigma_1 \) is the lower bound of the first Gaussian, and \( \mu_2 + \sigma_2 \) is the upper bound of the second Gaussian.
A higher $SepScore$ indicates greater separation.
The top-scoring image is chosen.
Finally, we define the sub-space associated with the concept. 
To achieve this, we focus on the Gaussian with the larger mean similarity,
as it indicates stronger similarity among the samples and therefore represents the concept. 
We denote this Gaussian as $\mathcal{N}_s(\mu_s,\sigma_s^2)$. 
For each embedding $e_j \in \mathcal{N}_s(\mu_s,\sigma_s^2)$ in the neighborhood, we compute its probability of belonging to this Gaussian as follows:
\begin{equation}
  Pr(\mathbf{e}_j ) = \frac{\pi_s \mathcal{N}_s(\mathbf{e}_j|\mu_s, \sigma_s^2)}{\pi_s \mathcal{N}_s(\mathbf{e}_j|\mu_s, \sigma_s^2) + (1-\pi_s) \mathcal{N}_{\bar{s}}(\mathbf{e}_j|\mu_{\bar{s}}, \sigma_{\bar{s}}^2)}.
\label{eq:prob}
\end{equation}
The concept subspace $\mathcal{C}(e, s,\tau)$, defined for an input embedding $e$ and surrogate $s$, consists of samples whose membership probabilities exceed the threshold $\tau$, that is, embeddings with higher similarity scores:
\begin{equation}
 \mathcal{C}(e, s,\tau) = \{\mathbf{e}_i | \mathbf{e}_i \in Neighbr_T(e), Prob(\mathbf{e}_i) > \tau \}.
 \label{eq:concept-set}
\end{equation}

\begin{figure}
    \centering
    \renewcommand{\arraystretch}{0.5}
    \setlength{\tabcolsep}{1pt}
    \begin{tabular}{c c c c}
        \multirow{6}{*}[-1.4em]{
            \centering
            \includegraphics[width=0.16\textwidth,keepaspectratio]{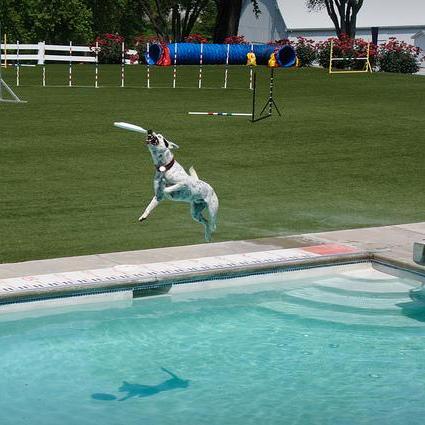}
        }
        & \includegraphics[width=0.1\textwidth,keepaspectratio]{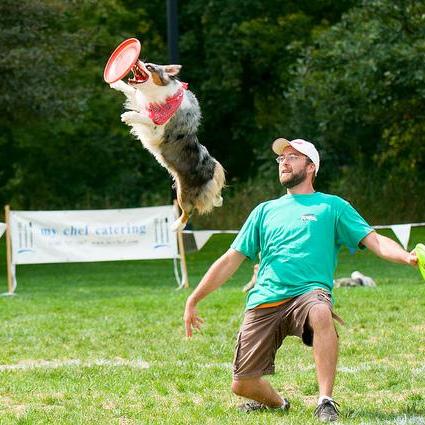}
        & \includegraphics[width=0.1\textwidth,keepaspectratio]{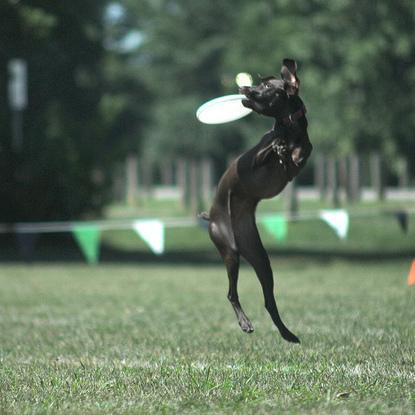}
        & \includegraphics[width=0.1\textwidth,keepaspectratio]{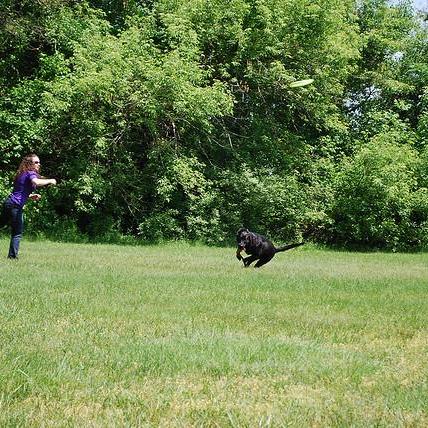} 
        \\
        & \includegraphics[width=0.1\textwidth,keepaspectratio]{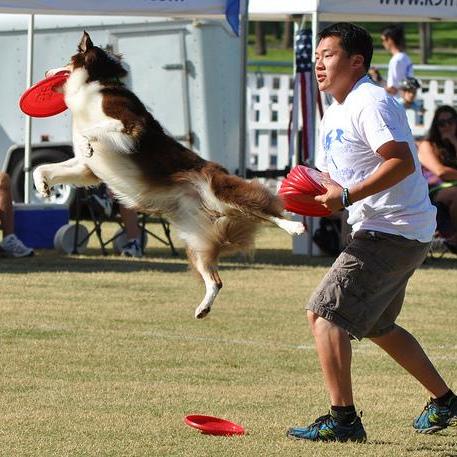}
        & \includegraphics[width=0.1\textwidth,keepaspectratio]{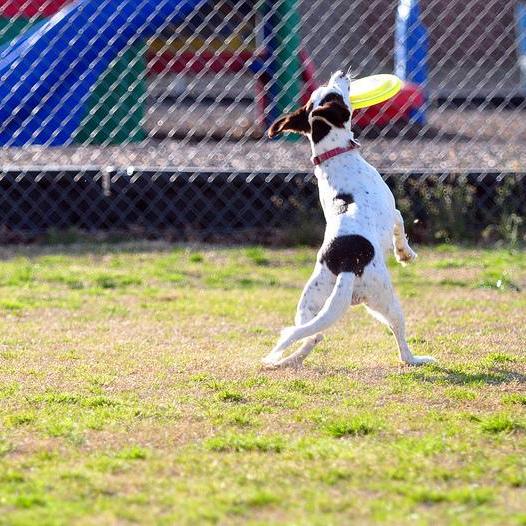}
        & \includegraphics[width=0.1\textwidth,keepaspectratio]{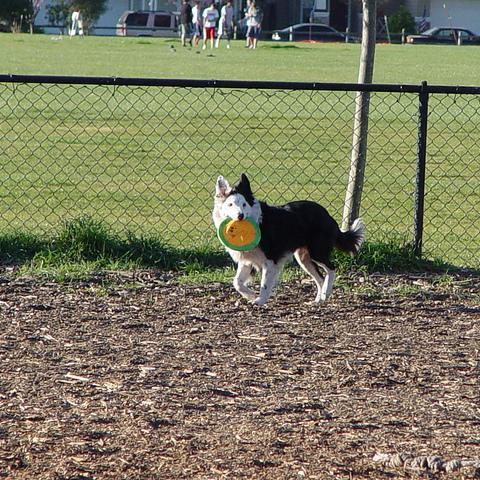} \\
        & \includegraphics[width=0.1\textwidth,keepaspectratio]{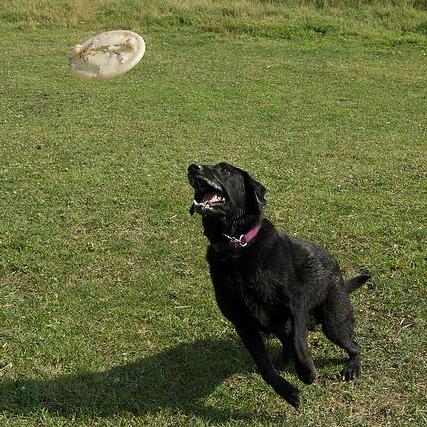}
        & \includegraphics[width=0.1\textwidth,keepaspectratio]{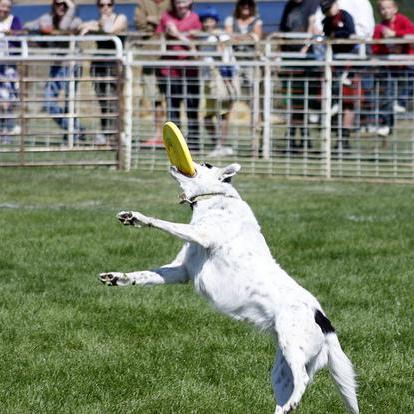}
        & \includegraphics[width=0.1\textwidth,keepaspectratio]{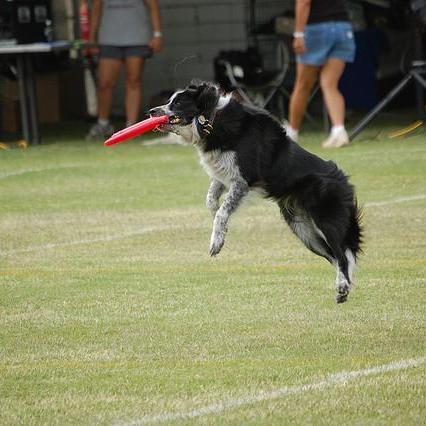} \\
        & \includegraphics[width=0.1\textwidth,keepaspectratio]{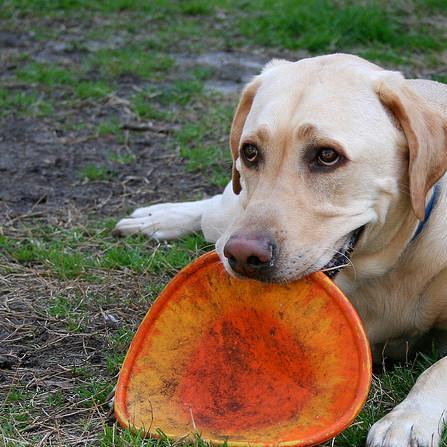}
        & \includegraphics[width=0.1\textwidth,keepaspectratio]{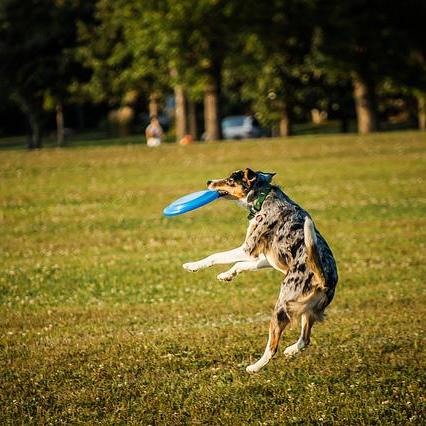}
        & \includegraphics[width=0.1\textwidth,keepaspectratio]{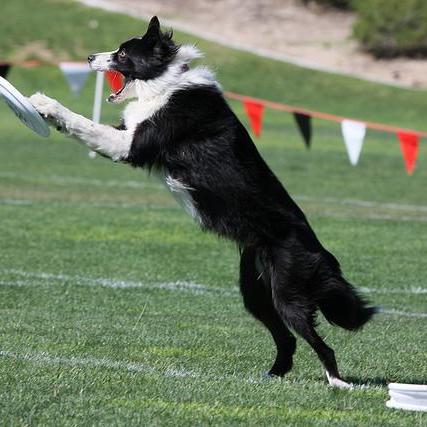}  \\
        Input & PCA 1 & PCA 2 & PCA 3 
    \end{tabular}
    \caption{
    {\bf PCA navigations.}
Each column represents a direction within the PCA subspace.
They display variations in (1) jump height, (2)~breed, and (3) distance from the camera.
However, they all share the underlying concept of 'a dog jumping for a frisbee.'
       }
    \label{fig:pca_control_grid_68729}
\end{figure}

\noindent
{\bf 3. Concept extraction.}
The objective of this step is to extract images that share a concept emerging from Eq.~\ref{eq:concept-set}. 
Although the set $\mathcal{C}(e, s, \tau)$ captures a single general concept, we may wish to focus on specific variations.
For example, each column in Fig.~\ref{fig:pca_control_grid_68729} illustrates a distinct variation of the concept 'a dog jumping for a frisbee.'
We propose a three-step process to extract a subset of  $\mathcal{C}(e, s, \tau)$:
(1)~constructing a subspace of the concept,
(2) projecting the input onto this subspace, and
(3)~extracting images corresponding to the concept by identifying the closest embeddings within this subspace.
We elaborate on these steps below.
To create a concept subspace, we aim to capture the main attributes of the concept. This is achieved by applying \textit{Principal Component Analysis (PCA)} to $\mathcal{C}(e, s, \tau)$, allowing us to extract the dominant components that define the concept.
The PCA-concept subspace is defined by the top $k$ principal components, corresponding to the largest singular values, and is spanned by these vectors, represented in the matrix $W_k \in \mathbb{R}^{d \times k}$. 
Fig.~\ref{fig:pca_control_grid_68729} demonstrates that navigating in different directions within the PCA space leads to variations of the concept.
PCA~1 varies with the dog's height, PCA~2 with breed, and PCA~3 with camera distance.
Sampling within the PCA subspace yields {\em inner-concept diversity.}
Up to this point, we have constructed a subspace that represents the concept, independent of the details of the input image.
To further enhance the relevance of the retrieved images to the input, we project the input embedding  \(\mathbf{e} \in \mathbb{R}^{d}\) onto the PCA-concept subspace. 
This isolates the attributes relevant to the concept, as the image may contain multiple overlapping concepts.
This is achieved by computing:
\begin{equation}
\mathbf{e}_{\text{c}} = \mathbf{e} \, \mathbf{W}_k \, \mathbf{W}_k^\top.
\label{eq:proj}
\end{equation}
Recall that the first principal component captures the largest possible variance, while subsequent ones capture progressively less.
The key question is how many components to select to capture the concept’s essence while minimizing noise.
The ratio between the sum of the variances of the top $k$ PCA components and the total variance quantifies how well the reduced representation preserves the original spread of the data.
Specifically, we use a captured variance threshold of  $\tau_1=25\%$, ensuring that the reduced representation remains meaningful and informative.
Thus, we choose the smallest $k$ that satisfies this criterion:
\begin{equation}
k = \min \left\{ k \mid \sum_{i=1}^{k} \frac{\sigma_i^2}{\sum_{j=1}^{d} \sigma_j^2} \geq \tau_1 \right\}
\label{eq:argmin-threshold}
\end{equation}
where $\sigma_i$ is the std of the $i^{th}$ principal component.
Finally, the  concept embedding $\mathbf{e}_{\text{c}}$ is used to retrieve the nearest neighbors. 
This is accomplished by applying cosine similarity to identify the most similar embeddings.
\vspace{0.02in}

\noindent
{\bf 4. Update the dataset and iterate.}
After processing the current concept, we shift focus to the next to promote {\em cross-concept diversity.}
The aim is to isolate concept-specific attributes and and reduce their influence in later iterations. 
This is achieved by subtracting the identified concept from a targeted subset of dataset embeddings, ensuring the next selected Gaussian captures a distinct concept.
The subtraction is carried out as follows: for each image embedding $e_i$ in the subset, we begin by computing its projection onto the concept subspace, as defined in Eq.~\ref{eq:proj}.
Next, we subtract this projection from the embedding of the corresponding image and update the dataset accordingly:
\begin{equation}
 \mathbf{e_i} \leftarrow \mathbf{e_i} - \mathbf{e_i} \, \mathbf{W}_k \, \mathbf{W}_k^\top.
    \label{eq:sub}
\end{equation}
The subset to which we apply Eq.~\ref{eq:sub} contains the top dataset embeddings that best match (Eq.~\ref{eq:sim}) the mean embedding of $\mathcal{C}(e,s,\tau)$.
We focus on a subset, not the full dataset, since the removed concept may co-occur with others.
Our aim is to avoid re-extracting it as a standalone concept, not to eliminate it entirely.
For example, the astronaut in Concept 1 of Fig.~\ref{fig:Teaser} may also appear in other concepts, like the cartoon in Concept 5.
This iterative process ensures that every concept 
is different from the previous ones, thereby allowing the exploration of multiple, potentially overlapping, concepts.
\section{Evaluation metrics}
\label{sec:evaluation}
There are no established metrics for our task.
Although related, image retrieval focuses on finding visually similar images, so standard metrics like precision and recall are not directly applicable.
Instead, evaluation should consider the four requirements in Section~\ref{sec:model}.
We propose a quantitative set of scores that is based on the concept embedding (Section~\ref{subsec:retrieval-metric})
Additionally, Section~\ref{subsec:human} also outlines our human study method.
\subsection{Quantitative metric}
\label{subsec:retrieval-metric}
\noindent\textbf{Relevance Score (RS).} 
This metric evaluates a concept’s relevance to the input image.
A straightforward approach would be to compute the similarity between the input embedding \(\mathbf{e} \in \mathbb{R}^d\) and the concept embedding \(\mathbf{e}_c \in \mathbb{R}^d\) from Eq.~\ref{eq:proj}, using \(\mathrm{Sim}(\mathbf{e}, \mathbf{e}_c)\) (Eq.~\ref{eq:sim}).  
But this raw similarity is too general, yielding retrieval-like results rather than capturing a specific concept.
To address this, we normalize similarity relative to concept distributions across the dataset.
The challenge is defining this normalization. 
We suggest extracting concepts for all images in the dataset (or approximating this using a random subset).  
Given this set, the similarity scores between the input image and the concepts form a Gaussian distribution with mean \(\mu\) and standard deviation~\(\sigma\).  
The normalized relevance is then defined as:
\begin{equation}
\mathrm{RS}(\mathbf{e}, \mathbf{e}_{c}) = \Phi\!\left(\frac{\mathrm{Sim}(\mathbf{e}, \mathbf{e}_{c})-\mu}{\sigma}\right),
\label{eq:rs_imrs}
\end{equation}
where $\Phi$ is the cumulative distribution function of the standard normal distribution.
This measures how the similarity score deviates from the overall distribution.  
The image-level relevance score, \(\mathrm{ImRS}(I)\), is defined as the sum of the relevance scores of all extracted concepts for image \(I\):
\begin{equation}
\mathrm{ImRS}(I) = \tfrac{1}{|\text{Conc}(\mathbf{e})|}\sum_{\mathbf{e}_{c} \in \text{Conc}(\mathbf{e})} \mathrm{RS}(\mathbf{e}_{c}).
\label{eq:rs_imrs}
\end{equation}
\noindent\textbf{Consistency Score (CS).} 
This metric measures the consistency of images retrieved within a concept.
Since the exact decomposition of an image into concepts is unknown, we compute the normalized sum of embeddings.
Unrelated concepts, being largely uncorrelated, cancel out when combined.
In contrast, a consistent concept shared across images reinforces certain entries, yielding a higher magnitude.
Formally, given \( n \) retrieved image embeddings \(\{{\mathbf{e}_j}\}_{j=1}^n\) for concept \(\mathbf{e}_{c}\), we define the concept-level consistency score \(\mathrm{CS}(\mathbf{e}_{c})\) and the image-level score \(\mathrm{ImCS}(I)\) as:
\begin{equation}
\begin{array}{rcl}
\mathrm{CS}(\mathbf{e}_{c}) &=& ||\tfrac{1}{n} \sum_{j=1}^{n} \mathbf{e}_j||, \quad \\
\mathrm{ImCS}(I) &=& \tfrac{1}{|\text{Conc}(\mathbf{e})|}\sum_{\mathbf{e}_{c} \in \text{Conc}(\mathbf{e})} \mathrm{CS}(\mathbf{e}_{c}).
\end{array}
\label{eq:cs_imcs}
\end{equation}
\begin{figure*}[t]
  \centering
    \begin{subfigure}[t]{0.48\textwidth}
    \centering
    \begin{subfigure}[c]{0.35\textwidth}
      \centering
      \includegraphics[width=\textwidth]{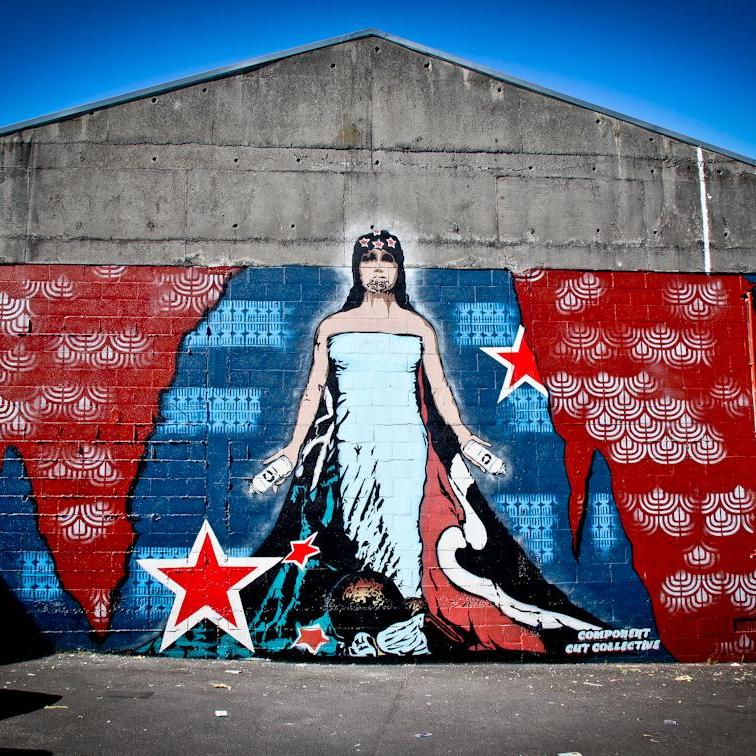}
      \caption*{\textbf{Input}}
    \end{subfigure}
    \hfill
    \begin{subfigure}[c]{0.60\textwidth}
      \centering
      \renewcommand{\arraystretch}{0.7}
      \setlength{\tabcolsep}{2pt}
      \begin{tabular}{ccc}
        \includegraphics[width=0.3\textwidth]{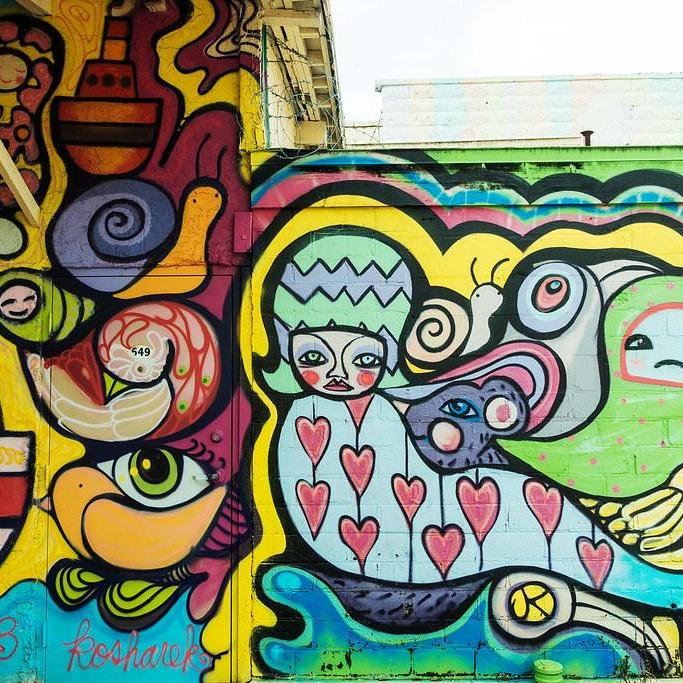} &
        \includegraphics[width=0.3\textwidth]{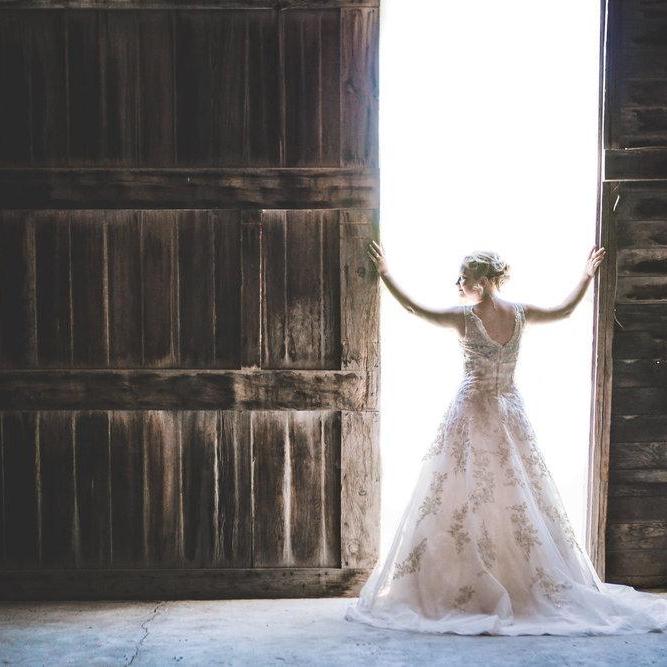} &
        \includegraphics[width=0.3\textwidth]{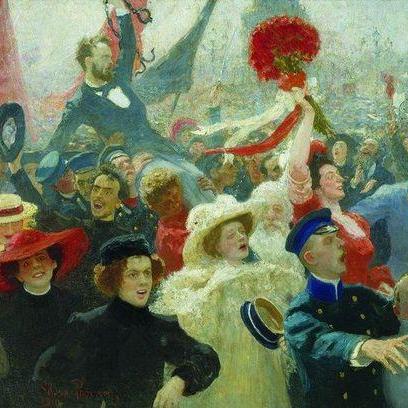} \\
        \includegraphics[width=0.3\textwidth]{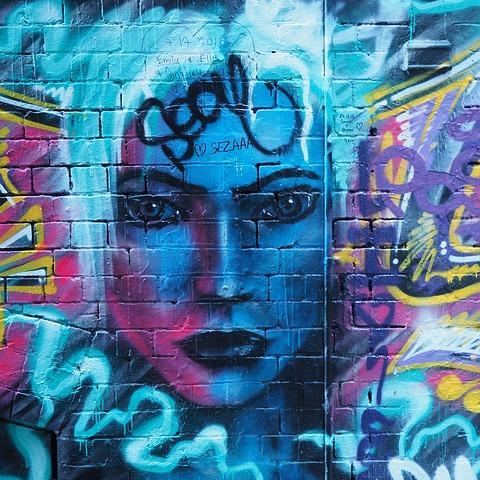} &
        \includegraphics[width=0.3\textwidth]{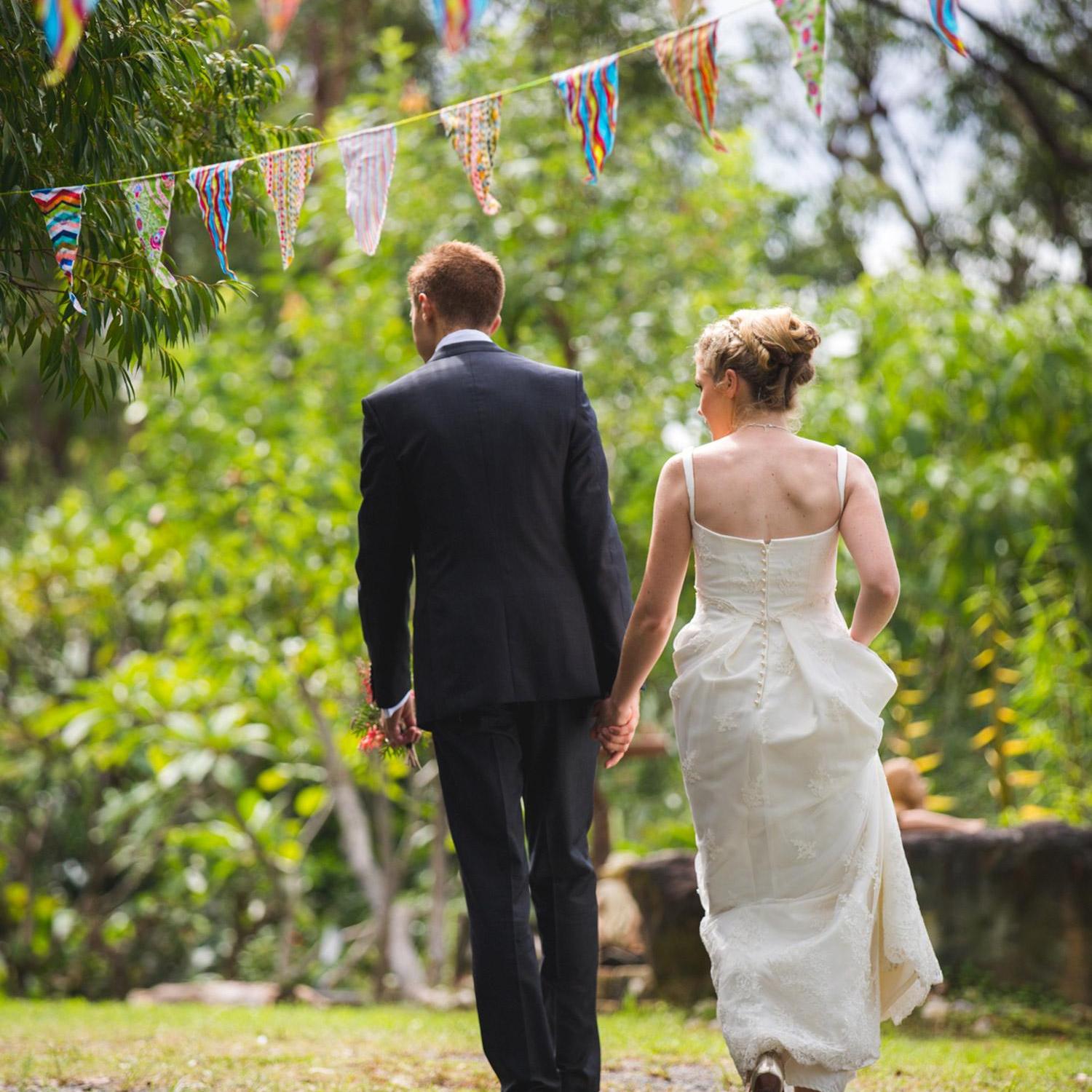} &
        \includegraphics[width=0.3\textwidth]{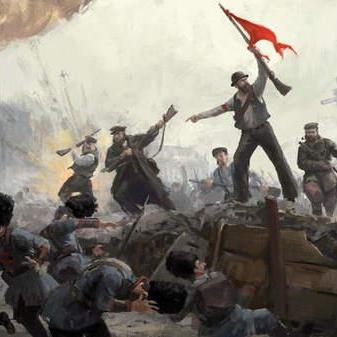} \\
        \includegraphics[width=0.3\textwidth]{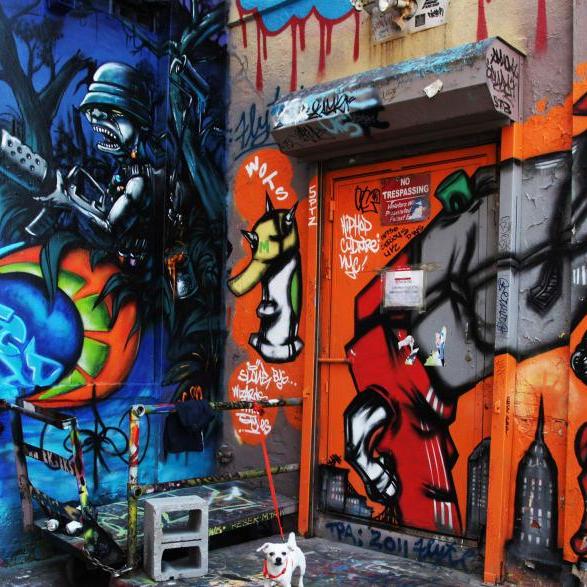} &
        \includegraphics[width=0.3\textwidth]{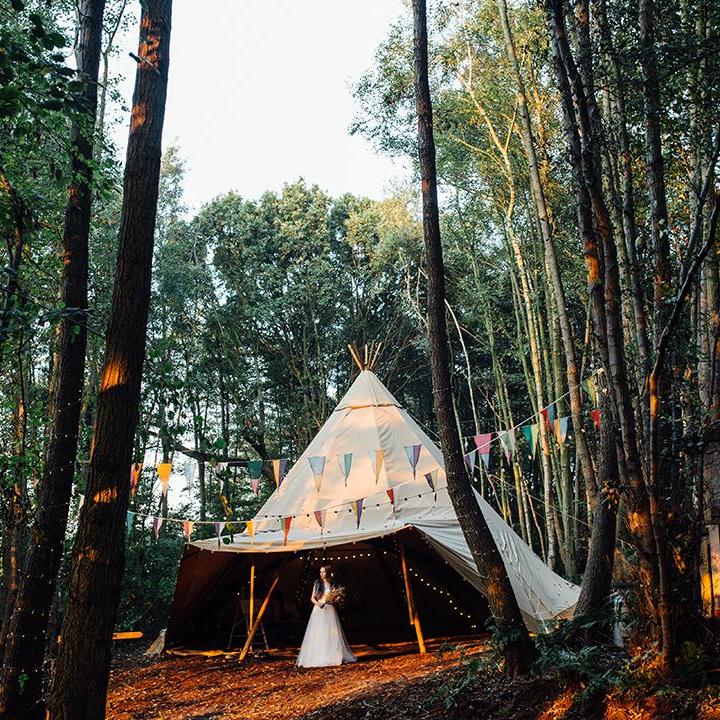} & 
        \includegraphics[width=0.3\textwidth]{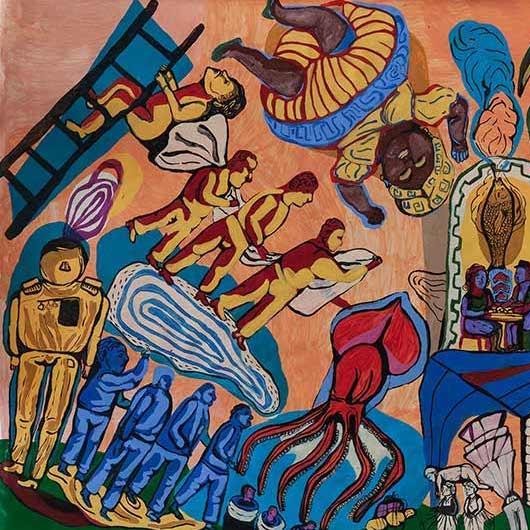} \\
        1 & 2 & 3 \\
      \end{tabular}
    \end{subfigure}
    \caption[.]{\raggedright
      The images retrieved by our algorithm can be interpreted as: 
      (1) 'Street art,' (2) 'Woman in white,' and (3) 'Revolution' (communist red star). 
    }
    \label{fig:combined_results_concept_extraction_sub_d}
  \end{subfigure}
  \hfill
  \begin{subfigure}[t]{0.48\textwidth}
    \centering
    \begin{subfigure}[c]{0.35\textwidth}
      \centering
      \includegraphics[width=\textwidth]{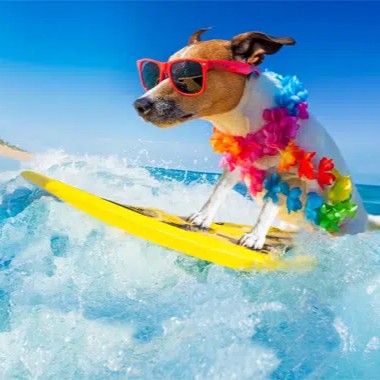}
      \caption*{\textbf{Input}}
    \end{subfigure}
    \hfill
    \begin{subfigure}[c]{0.60\textwidth}
      \centering
      \renewcommand{\arraystretch}{0.7}
      \setlength{\tabcolsep}{2pt}
      \begin{tabular}{ccc}
        \includegraphics[width=0.3\textwidth]{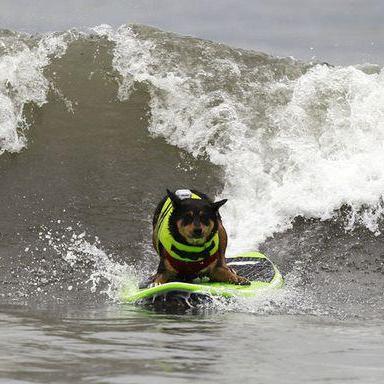} &
        \includegraphics[width=0.3\textwidth]{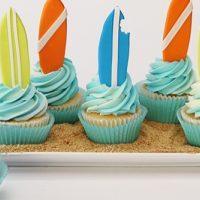} &
        \includegraphics[width=0.3\textwidth]{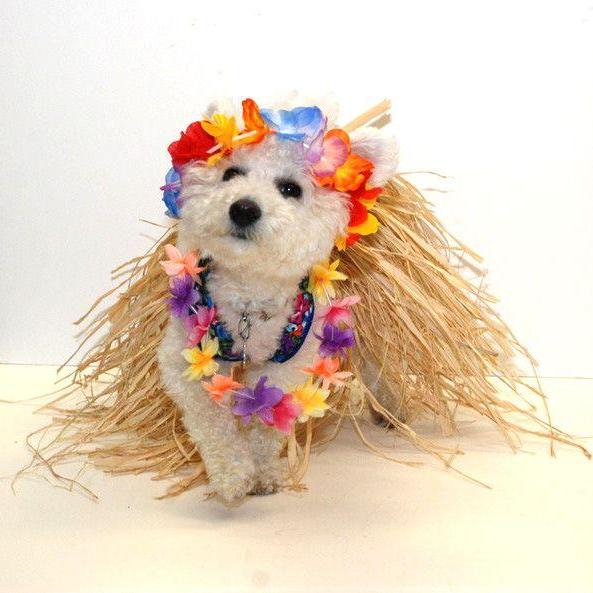} \\
        \includegraphics[width=0.3\textwidth]{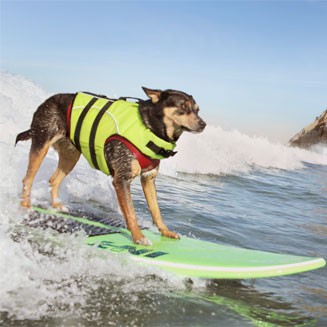} &
        \includegraphics[width=0.3\textwidth]{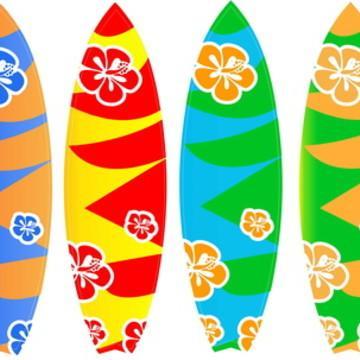} &
        \includegraphics[width=0.3\textwidth]{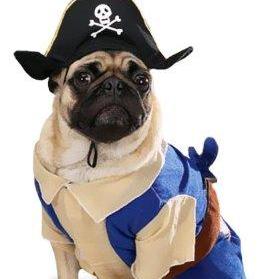}\\
        \includegraphics[width=0.3\textwidth]{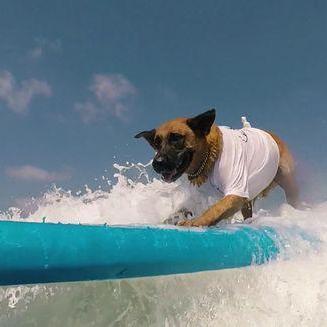}  &
        \includegraphics[width=0.3\textwidth]{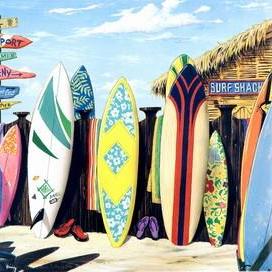} &
        \includegraphics[width=0.3\textwidth]{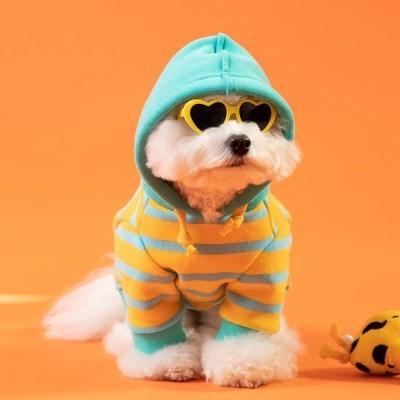} \\
        1 & 2 & 3 \\
      \end{tabular}
    \end{subfigure}
    \caption[.]{\raggedright
      The images retrieved by our algorithm can be interpreted as: 
      (1) 'A surfing dog' (2) 'Surfing boards,' 
      and (3) 'A dog in funny costume.'
    }
    \label{fig:combined_results_concept_extraction_sub_b}
  \end{subfigure}
  \vskip 1em
  \begin{subfigure}[t]{0.48\textwidth}
    \centering
    \begin{subfigure}[c]{0.35\textwidth}
      \centering
      \includegraphics[width=\textwidth]{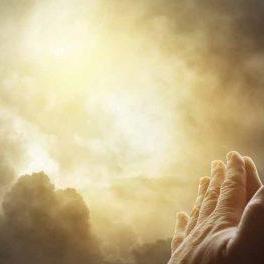}
      \caption*{\textbf{Input}}
    \end{subfigure}
    \hfill
    \begin{subfigure}[c]{0.60\textwidth}
      \centering
      \renewcommand{\arraystretch}{0.7}
      \setlength{\tabcolsep}{2pt}
      \begin{tabular}{ccc}
        \includegraphics[width=0.3\textwidth]{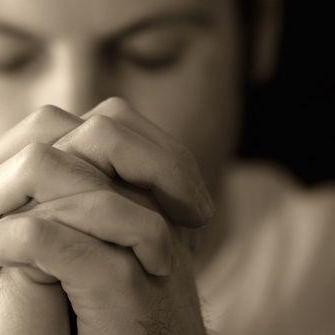} &
        \includegraphics[width=0.3\textwidth]{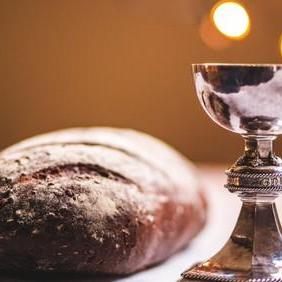} &
        \includegraphics[width=0.3\textwidth]{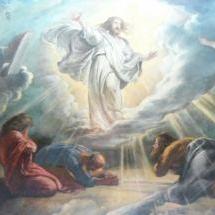} \\
        \includegraphics[width=0.3\textwidth]{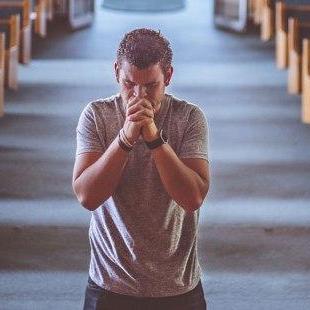} &
        \includegraphics[width=0.3\textwidth]{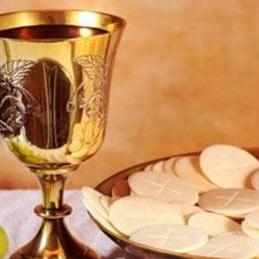} &
        \includegraphics[width=0.3\textwidth]{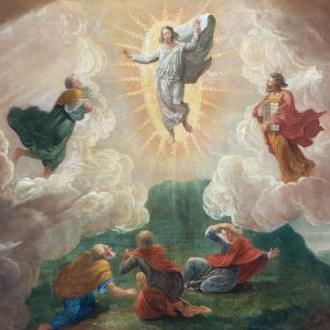} \\
        \includegraphics[width=0.3\textwidth]{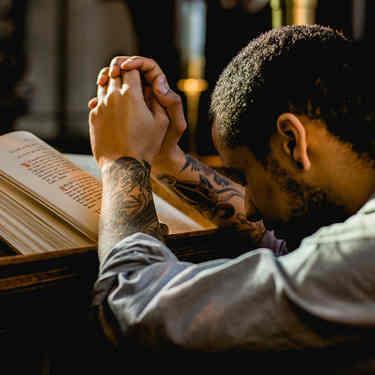} &
        \includegraphics[width=0.3\textwidth]{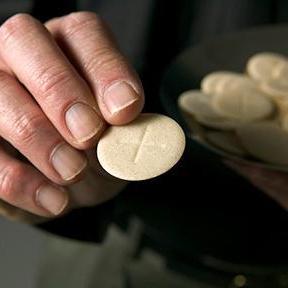} &
        \includegraphics[width=0.3\textwidth]{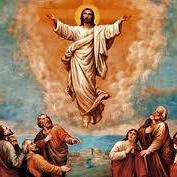} \\
        1 & 2 & 3 \\
      \end{tabular}
    \end{subfigure}
    \caption[.]{\raggedright
      The images retrieved by our algorithm can be interpreted as: (1) 'A praying person,'
      (2) 'Sacramental bread,' and (3) 'Heavenly ascension.'
    }
    \label{fig:combined_results_concept_extraction_sub_a}
  \end{subfigure}
  \hfill
    \begin{subfigure}[t]{0.48\textwidth}
    \centering
    \begin{subfigure}[c]{0.35\textwidth}
      \centering
      \includegraphics[width=\textwidth]{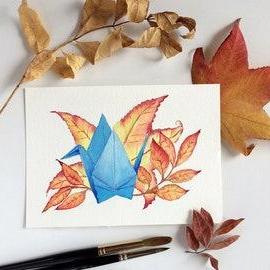}
      \caption*{\textbf{Input}}
    \end{subfigure}
    \hfill
    \begin{subfigure}[c]{0.60\textwidth}
      \centering
      \renewcommand{\arraystretch}{0.7}
      \setlength{\tabcolsep}{2pt}
      \begin{tabular}{ccc}
        \includegraphics[width=0.3\textwidth]{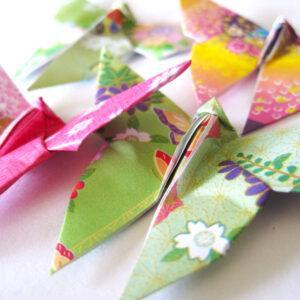} & 
        \includegraphics[width=0.3\textwidth]{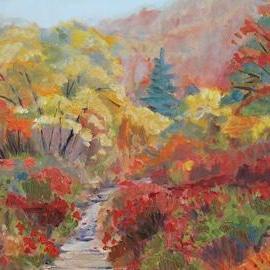} & 
        \includegraphics[width=0.3\textwidth]{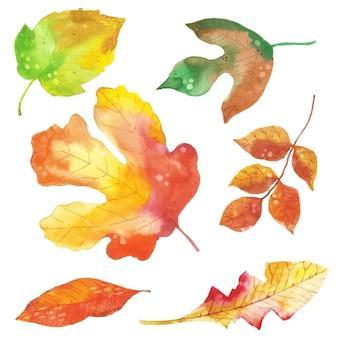} \\
        \includegraphics[width=0.3\textwidth]{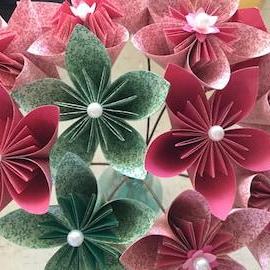} & 
        \includegraphics[width=0.3\textwidth]{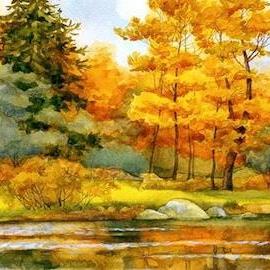} & 
        \includegraphics[width=0.3\textwidth]{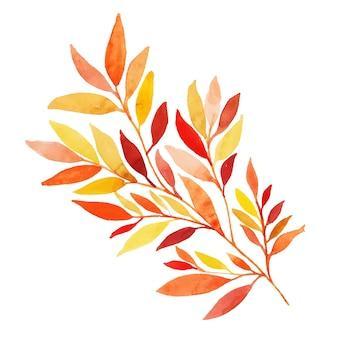} \\
        \includegraphics[width=0.3\textwidth]{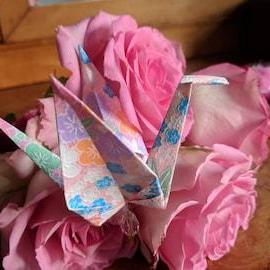} & 
        \includegraphics[width=0.3\textwidth]{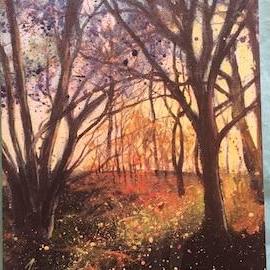} & 
        \includegraphics[width=0.3\textwidth]{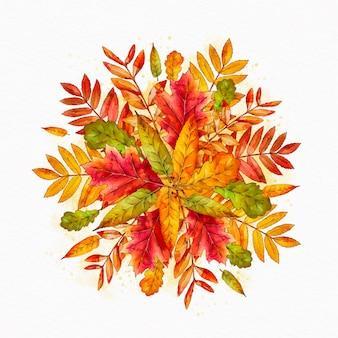} \\
        1 & 2 & 3 \\
    \end{tabular}
    \end{subfigure}
    \caption[.]{\raggedright
      The images retrieved by our algorithm can be interpreted as: 
      (1)~'Origami,' (2) 'Autumn,' and (3)~'A drawing of autumn leaves.'
    }
    \label{fig:combined_results_concept_extraction_sub_c}
  \end{subfigure}
    \vspace{8pt}
  \caption[.]{\raggedright
    {\bf Qualitative results.}
Each subfigure shows an input image with three columns, each depicting a distinct extracted concept.
All concepts are relevant, consistent, diverse and quite creative.
See supplementary material for additional examples.
  }
  \label{fig:combined_results_concept_extraction}
\end{figure*}
\noindent\textbf{Inner-Diversity Score (IDS).}
This metric quantifies variance among retrieved images for a concept, aiming to span a subspace that captures it.
As shown in Fig.~\ref{fig:pca_control_grid_68729}, variations arise by exploring the subspace in different directions.
Therefore, we approximate this subspace with Principal Component Analysis (PCA) on the embeddings of retrieved images.
Diversity is then measured as the cumulative variance explained by the top $K$ principal components.
Formally, given a concept embedding \( \mathbf{e}_{c} \) and its retrieved image embeddings,
we fit a PCA and obtain eigenvalues $\sigma^2_{1} \ge \dots \ge \sigma^2_{d}$.
With $K$ leading components retained, the concept-level $\mathrm{IDS}(\mathbf{e}_{c})$ and image-level $\mathrm{ImIDS}(I)$ inner-diversity scores are defined as:
\begin{equation}
\begin{array}{rcl}
\mathrm{IDS}(\mathbf{e}_{c}) &=&
\nicefrac{\displaystyle\sum_{k=1}^{K}\sigma^2_k}
     {\displaystyle\sum_{k=1}^{d}\sigma^2_k}, \quad \\
\mathrm{ImIDS}(I) &=& \frac{1}{|\text{Conc}(\mathbf{e})|}\sum_{\mathbf{e}_{c}\in\text{Conc}(\mathbf{e})}\mathrm{IDS}(\mathbf{e}_{c}),
\end{array}
\label{eq:ids_imids}
\end{equation}
where a larger value indicates higher inner diversity.
\noindent\textbf{Cross-Diversity Score (CDS).} 
This metric captures variance across concepts.
To promote diversity, we first quantify their pairwise distinctiveness as:
\[
\mathrm{CDS}(\mathbf{e}_{c_i}, \mathbf{e}_{c_j}) = \tfrac{1}{2} \bigl(1 - \mathrm{Sim}(\mathbf{e}_{c_i}, \mathbf{e}_{c_j})\bigl),
\]
where higher values indicate greater diversity.
We define the concept-level $\mathrm{conceptCDS}(\mathbf{e}_{c})$ as the minimum CDS to other concepts, and the image-level $\mathrm{ImCDS}(I)$ as the normalized sum across concepts:
\begin{equation}
\begin{array}{rcl}
\mathrm{concCDS}(\mathbf{e}_{c}) &=& \min_{{\substack{ \mathbf{e}_{c_j} \neq \mathbf{e}_{c_i}}}} \mathrm{CDS}(\mathbf{e}_{c_i}, \mathbf{e}_{c_j}), \quad \\
\mathrm{ImCDS}(I) &=& \tfrac{1}{|\text{Conc}(\mathbf{e})|} \sum_{\mathbf{e}_{c} \in \text{Conc}(\mathbf{e})} \mathrm{concCDS}(\mathbf{e}_{c}).
\end{array}
\label{eq:imcds}
\end{equation}
\subsection{Human evaluation methodology}
\label{subsec:human}
Since visual concepts are inherently abstract, quantitative metrics may not fully capture human intent.
We therefore propose a human evaluation to assess the quality of retrieved concepts against the key requirements.

\noindent
{\bf Evaluating consistency.}
First, participants are asked to evaluate the extent to which a given set of images share a common concept.
Next, they are instructed to describe the identified concept in 1–4 words of free text.
This setup was designed to test whether the retrieved images convey a consistent concept on their own.
Importantly, at this stage participants do not have access to the input image.

\noindent
{\bf Evaluating relevance.}
Next, a source image is shown to participants, who must determine whether their identified concept is fully contained, partially contained (or related), or not contained in the source image.
The source image may be the original input, a randomly selected image, or one with the same objects arranged differently.
The random image is expected to show no relevance, the re-arranged image partial relevance, and the input image to share the concept.
Thus, the first two serve as negative controls.

\noindent
{\bf Evaluating inner-concept diversity.}
Participants are presented with the input image alongside two retrieved sets: one generated by our concept-retrieval algorithm and the other by classical retrieval~\cite{dfn_fang2023data, radford2021clip}.
They are then asked to compare their diversity by choosing one of the following:
(1) Both sets exhibit similar diversity.
(2) The first set (concept-retrieval) is more diverse.
(3) The second set (classical retrieval) is more diverse.

\noindent
{\bf Evaluating cross-concept diversity.}
To assess concept distinctiveness, participants are shown the input image with three related concept sets.
For each set, they first repeat the relevance evaluation and then choose one of the following:
(1) Yes, the concepts are different.
(2) No, the concepts are not different.
(3) Similar to one but different from another.
\section{Results}
\label{sec:results}

\noindent
{\bf Datasets. }
To evaluate the proposed method, we used diverse datasets spanning different domains and semantic complexities.  
\emph{COCO}~\cite{lin2014microsoft_coco_dataset} contains \(330{,}000\) images with 80 object categories across various contexts.
{\em LAION-Aesthetics} is a subset of the LAION-5B~\cite{schuhmann2022laionb} dataset, curated to include images of high aesthetic quality.
It comprises millions of images rated based on aesthetic scores, capturing diverse visual styles.
{\em DeepFashion}~\cite{liuLQWTcvpr16DeepFashion} consists of over $800,000$ diverse fashion images, each annotated with $50$ clothing categories and $1,000$ descriptive attributes.

\vspace{0.02in}
\noindent
{\bf Qualitative results.}
Figure~\ref{fig:combined_results_concept_extraction} shows some examples of our algorithm applied to images from~\cite{lin2014microsoft_coco_dataset,schuhmann2022laionb}. 
The input images span diverse contexts, including urban artistic expressions, animals and sports activities, religious scenes, and nature in art.
For instance, in Figure~\ref{fig:combined_results_concept_extraction_sub_d},
our method extracts three sets of images, whose shared concepts may be interpreted as:
(1) 'street art,'
(2) 'a woman wearing a long, white dress,'
and (3) 'revolution (an arm projecting a sense of power, with political/social symbolism such as a red star, flag, flowers, or stone).'
The retrieved images satisfy the four key requirements.
They are relevant, containing visual elements from the input image. Each set maintains a consistent theme. 
Inner-diversity is reflected in variations of pose and scene details. 
Finally, the three concept sets remain visually and semantically distinct, ensuring cross-diversity.
\begin{figure}
  \centering
  \begin{subfigure}[c]{0.165\textwidth}
    \centering
    \includegraphics[width=\textwidth]{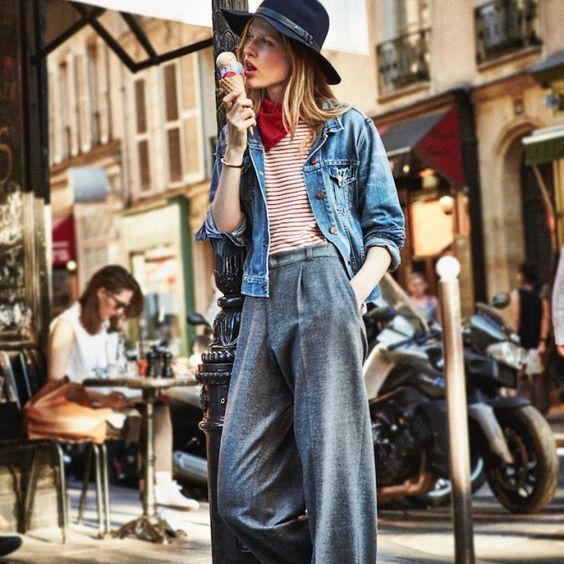}
    \caption*{\textbf{Input}}
  \end{subfigure}
  \hfill
  \begin{subfigure}[c]{0.30\textwidth}
    \centering
    \renewcommand{\arraystretch}{0.7}
    \setlength{\tabcolsep}{2pt}
    \begin{tabular}{ccc}
      \includegraphics[width=0.3\textwidth]{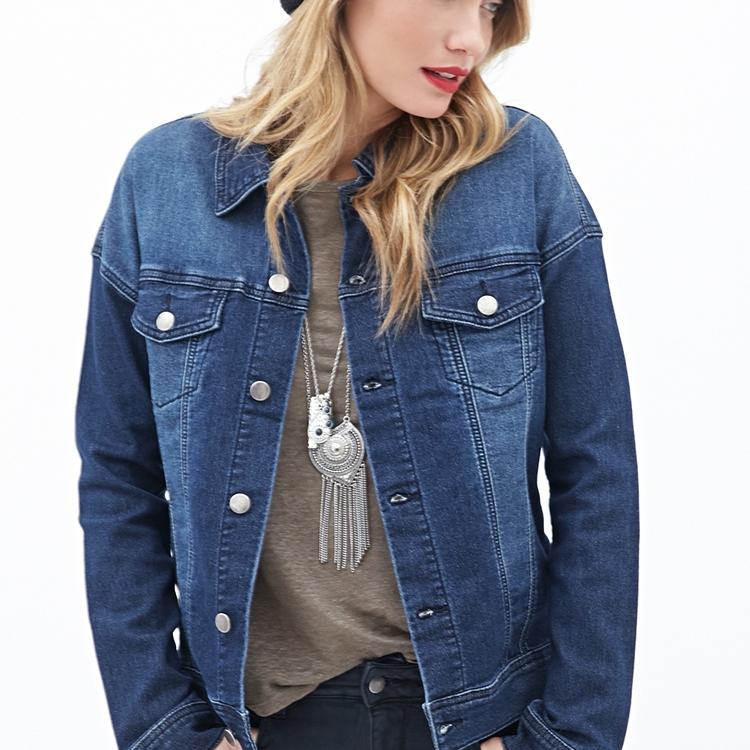} &
      \includegraphics[width=0.3\textwidth]{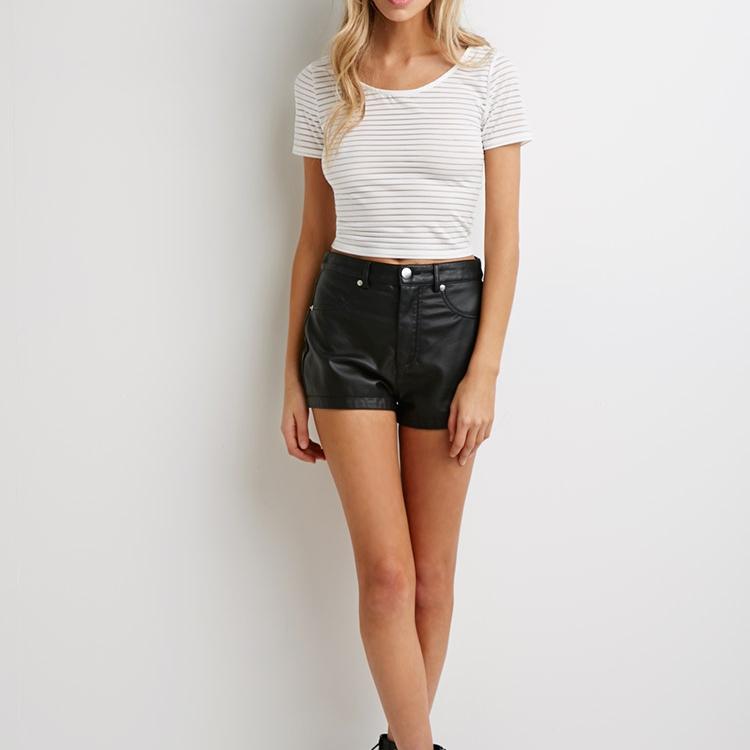} &
      \includegraphics[width=0.3\textwidth]{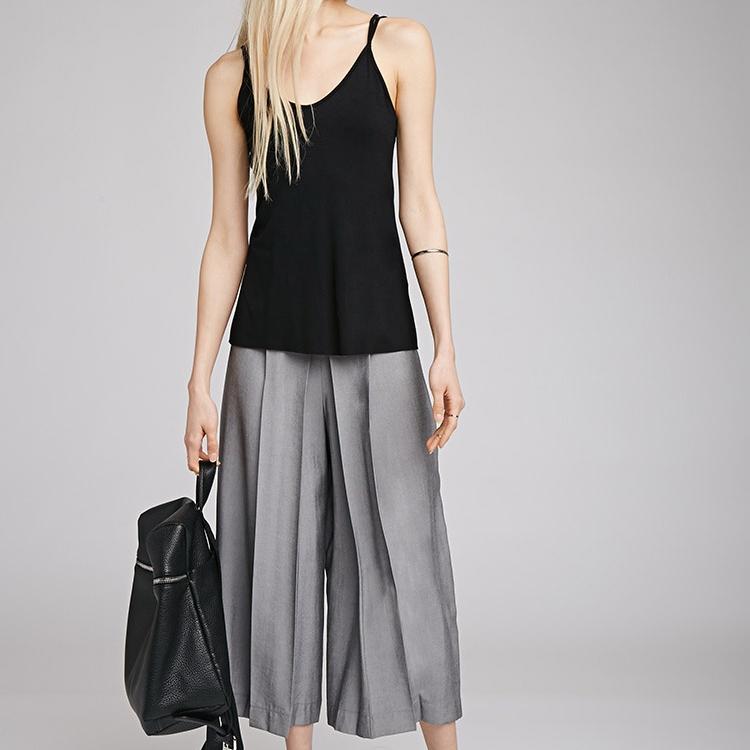} \\
      \includegraphics[width=0.3\textwidth]{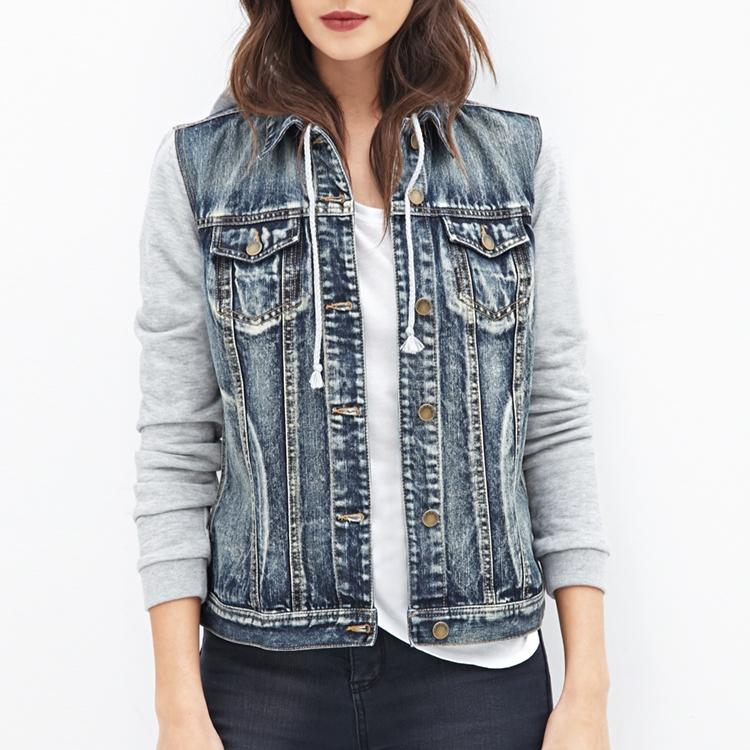} &
      \includegraphics[width=0.3\textwidth]{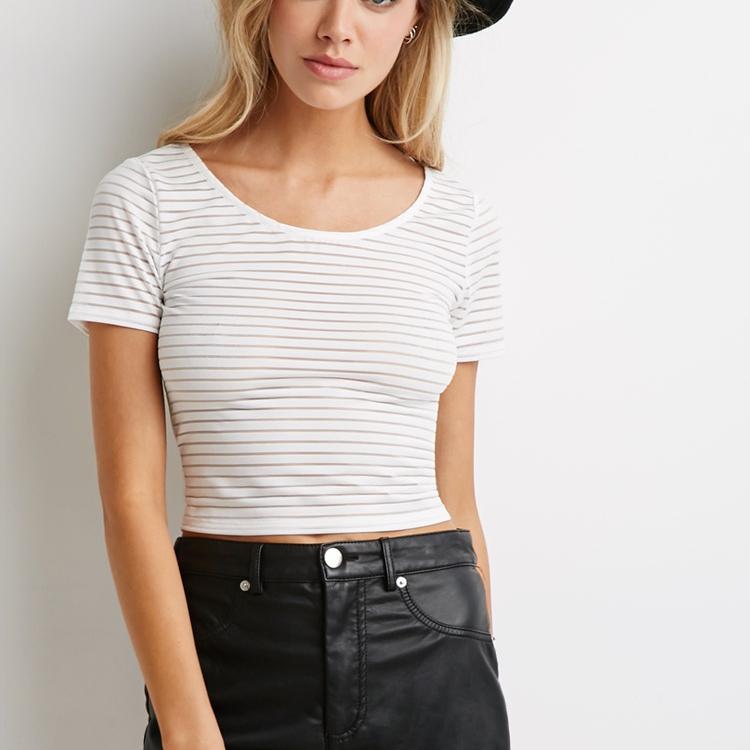} &
      \includegraphics[width=0.3\textwidth]{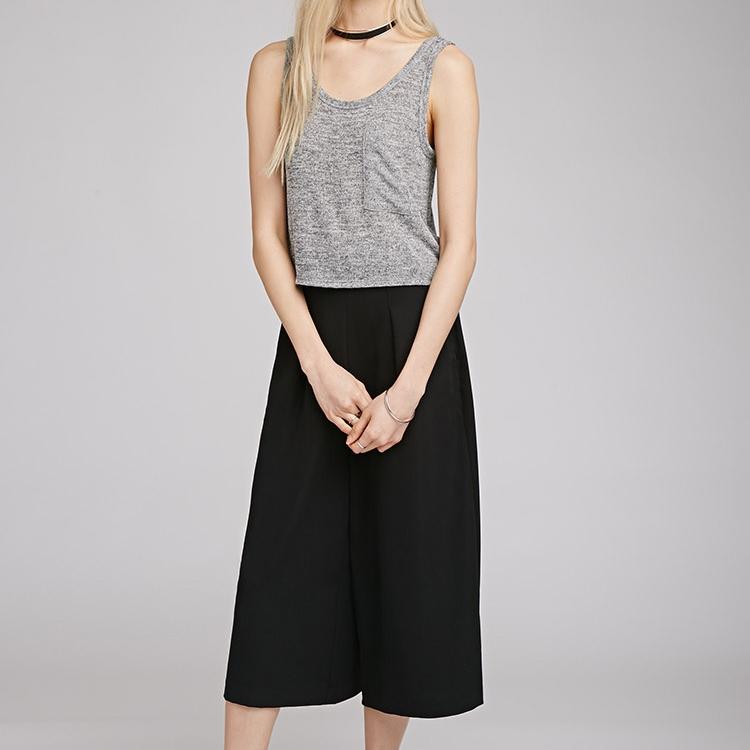} \\
      \includegraphics[width=0.3\textwidth]{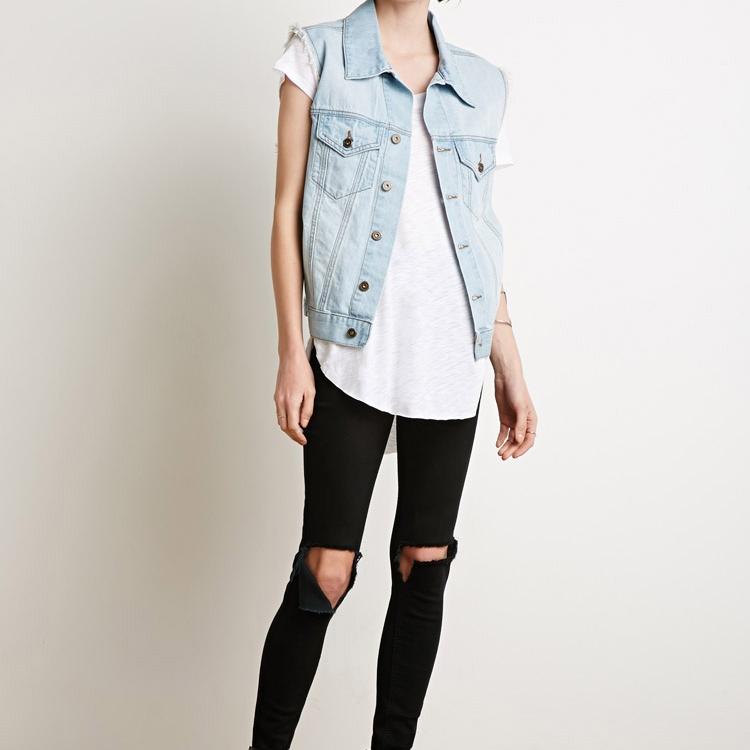} &
      \includegraphics[width=0.3\textwidth]{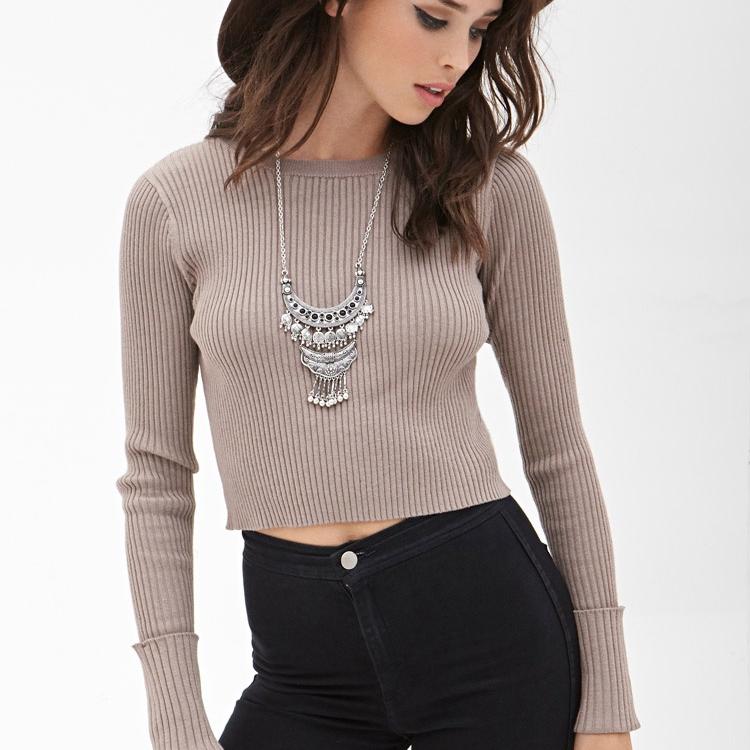} &
      \includegraphics[width=0.3\textwidth]{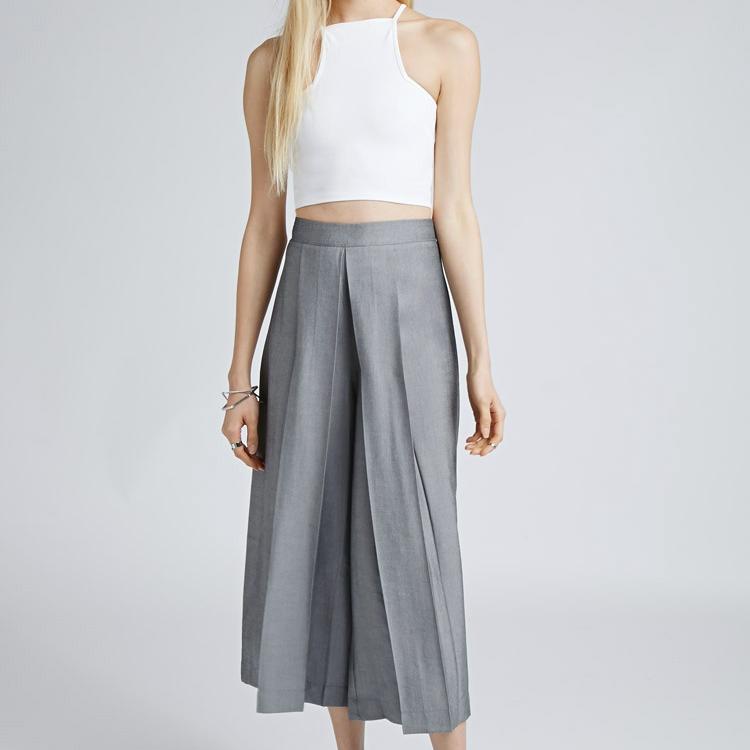} \\
      1 & 2 & 3
    \end{tabular}
  \end{subfigure}
    \caption{
    \textbf{Domain‐specific concepts.} 
    }
  \label{fig:aesthetic_concepts_wrap}
\end{figure}
Figure~\ref{fig:aesthetic_concepts_wrap} demonstrates a domain-specific result on an input image from  LAION-Aesthetics \cite{schuhmann2022laionb}, using the Deep Fashion dataset~\cite{liuLQWTcvpr16DeepFashion} for concept extraction.
Our method efficiently decomposes the input image into relevant domain-specific concepts, such as 'women in denim outerwear,' 'striped tops,' and 'wide-leg grey bottoms.'
This capability has applications in fashion recommendation systems, virtual try-on solutions, and personalized shopping experiences, enhancing user interaction and retrieval accuracy in fashion-related tasks.
Additional general and domain-specific results are in the supplementary material.
These examples highlight how our problem and results differ from standard image retrieval.
Instead of producing a single ranked list of globally similar images, the input image is implicitly decomposed into more abstract concepts.

\vspace{0.02in}
\noindent
{\bf Quantitative results.}
We compare our results to two baselines: (1) the retrieval method of~\cite{dfn_fang2023data, radford2021clip}, which retrieves 60 images divided into three sets, and (2) $k$-means clustering ($k=3$) in the embedding space, as a potential alternative for concept extraction.
\begin{table}
\centering
\small
\setlength{\tabcolsep}{4pt}
\begin{tabular}{lcccc}
\toprule
Method & $\text{ImRS}_{\mu}$ & $\text{ImCS}_{\mu}$ & $\text{ImIDS}_{\mu}$ & $\text{ImCDS}_{\mu}$ \\
\midrule
Ours      & 0.92      & 0.87      & {\bf0.59} & {\bf0.15} \\
K\_Means  & 0.98      & {\bf0.88} & 0.56      & 0.05 \\
Retrieval & {\bf0.99} & 0.83      & 0.54      & 0.01 \\
\bottomrule
\end{tabular}
\caption{
{\bf Quantitative results.}
As desired, our results show significantly greater diversity.
}
\label{tab:quantitative_1}
\end{table}
Table~\ref{tab:quantitative_1} shows our quantitative results averaged over all images in the diverse LAION-Aesthetics~\cite{schuhmann2022laionb} dataset.
Our method achieves the desired outcome: significantly higher diversity, both within and across concepts, while maintaining high relevance and consistency.
This is because we aim at extracting images that share conceptual similarities, regardless of other visual elements, thereby ensuring diversity.
As expected, the retrieval-based approach attains the highest Relevance Score, as it retrieves the most visually similar images, effectively capturing overall content rather than a diverse range of concepts.
Meanwhile, $k$-means clustering ranks highest in consistency, as it selects the closest images within a small neighborhood.
These results align with the human evaluation described next.

\vspace{0.02in}
\noindent
{\bf Human evaluation.}
We conducted a user study, as detailed in Section~\ref{subsec:human}, to assess whether human intuition regarding concepts
aligns with the concepts identified by our method.
A total of $32$ participants took part in the evaluation, including $15$ females and $17$ males.
The participants' ages ranged from $18$ to $75$ years.
In this study, participants were presented with $21$ sets of concepts derived from seven different input images from COCO~\cite{lin2014microsoft_coco_dataset}.
\noindent
{\em Evaluating consistency.}
Given sets of images extracted by our algorithm, each representing a concept, $95\%$ of participants recognized the images within each set as sharing a common concept. This result highlights the effectiveness of our algorithm in identifying meaningful concepts.
\noindent
{\em Evaluating relevance.}
Now, when presented with the input image, $79\%$ of the participants agreed that the concept represented by the set is indeed present in the input image ($62\%$ fully contained, $17\%$ partially contained).
We compared this relevance to two baseline methods:
(1)~A random input image, where only $33\%$ of participants found the concept relevant.
(2)~A retrieval algorithm based on object matching, where a subset of objects from the input image was selected, and images containing the same objects were retrieved.
In this case, only  $41\%$ of participants found the retrieved images relevant.
These results confirm that a concept is more than just a collection of objects.
\noindent
{\em Evaluating inner-concept diversity.}
When presented with the input image alongside two retrieved sets—one from our concept-retrieval algorithm and one from classical retrieval---our algorithm outperforms the baseline by $14\%$,
reflecting participants’ relative preference for our approach over retrieval baselines.
\noindent
{\em Evaluating cross-concept diversity.}
When shown the input along with three sets of related concepts, $90\%$ of participants agreed the sets represent different concepts. Among them, $67\%$ stated all three sets were completely different, while $23\%$ found two out of the three were different.
These results confirm that our approach extracts relevant, consistent, and internally and externally diverse concepts.
They also clearly demonstrate that a concept is more than the sum of its objects.

\vspace{0.02in}
\noindent
{\bf Ablation: key thresholds.}
Our algorithm uses several parameters, including the neighborhood size $T$, the reduced representation threshold $\tau$, and the percentage of embeddings modified between iterations, with detailed tests provided in the supplementary material.
Overall, our method is stable across a broad range of hyperparameter values. 
The chosen defaults provide a good balance between relevance, consistency, and both forms of diversity.
For update percentage, as expected larger updates reduce relevance and consistency but increase diversity.
Higher $\tau$ improves relevance but slightly reduces diversity and consistency.
Thus, the parameters were empirically selected: the neighborhood size was set to  \(T=0.25\sigma\) (Eq.~\ref{eq:neighbr}),
the reduced representation threshold was set to \(\tau_1=0.25\) (Eq.~\ref{eq:argmin-threshold}),
and the percentage of embeddings modified between iterations was $10\%$.

\vspace{0.02in}
\noindent
{\bf Generalization.}
The results presented in this paper utilize embeddings from ViT-H-14-DFF~\cite{dfn_fang2023data,radford2021clip}, due to its strong zero-shot capabilities and the rich semantic structure of its embedding space.
Our approach generalizes well across different vision models.
In addition to ViT-H-14-DFN, we tested it with weaker embeddings from CLIP ViT-L/14 and DINO.
As expected, retrieved concept quality decreases with weaker embeddings, as these models capture less semantic structure.Nevertheless, the method performs well across models, as demonstrated in the supplementary material.

\vspace{0.02in}
\noindent
{\bf Computational Efficiency:} 
Our method relies on repeated nearest-neighbor searches, Gaussian Mixture Model (GMM) fitting, and PCA computations.
Approximate nearest neighbor search has complexity \(O(n)\)~\cite{bentley1975multidimensional}, where \(n\) is the number of dataset images.  
Approximate PCA has complexity \(O(nd)\)~\cite{halko2011finding}, where \(n\) is a subset of the dataset, \(d\) is the embedding dimension, and the number of principal components is small.  
Approximate GMM~\cite{dempster1977maximum} has complexity \(O(n)\), where \(n\) again refers to a subset.
The overall complexity is thus bounded by \(O(nd)\).
The running time is $4.29$ seconds per image when extracting three concepts on an AMD EPYC 7763 CPU.

\vspace{0.02in}
\noindent
{\bf Limitations.}
Our approach lacks user control over the extracted concepts, which may 
particularly important for domain-specific applications.
This is an intriguing direction for future research.
Furthermore, the method might struggle when the concept is extremely rare in the dataset, as there may not be enough
 supporting samples to reliably separate it as a distinct mode.
\section{Conclusion}
\label{sec:conclusion}
This paper introduces the problem of image concept retrieval, which can be seen as a generalization of traditional image retrieval. 
We define the essential requirements for any method addressing this task and propose a novel approach for concept extraction based on these requirements. Our approach is built on two key observations: (1) the similarity distribution within the input's neighborhood can be modeled as a bimodal Gaussian, and (2) certain neighbors clearly exhibit this structure.
Our approach requires no training, which is crucial given the difficulty of obtaining ground truth.
Additionally, we introduce new evaluation metrics tailored to this task. Both qualitative and quantitative results, supported by a human study, validate the effectiveness of our approach.
As this is a new problem, several future directions are possible.
First, incorporating control over which concepts are extracted.
Second, extending our concept retrieval method to other embedding spaces, such as sentences or documents. Since the method relies on neighborhood similarity distributions, it should transfer naturally to such domains.
Finally, we expect that dedicated datasets for this task will further support community benchmarking.

\vspace{0.02in}
\noindent
{\bf Acknowledgments.} 
We gratefully acknowledge the support of the Israel Science Foundation (ISF) 2329/22.
\clearpage
{\small
\bibliographystyle{ieeenat_fullname}
\bibliography{main_arXiv}
}
\end{document}